%% file: main-submission-arxiv.tex
\documentclass[9.5pt,twocolumn,twoside]{IEEEtran}
\pdfoutput=1
\usepackage{graphicx}
\usepackage{amsmath,amsfonts,amssymb,amscd,amsthm,xspace}
\usepackage{import}
\usepackage{xcolor}
\usepackage{subfig}
\usepackage{mathtools}
\usepackage{math}
\usepackage{enumitem}
\usepackage[font=small,labelfont=bf,textfont=bf]{caption}
\usepackage{multirow}
\usepackage{booktabs}
\usepackage[noadjust,sort]{cite} % sort
\usepackage{url}
\usepackage{tikz}
\usepackage{relsize}
\usetikzlibrary{positioning}
\tikzset{fontscale/.style = {font=\relsize{#1}}}
\usepackage{pbox}

\usepackage[final]{review}
\setrevision{1}
\graphicspath{{../}}

\newcommand{\fref}[1]{Fig.~\ref{#1}}
\newcommand{\tref}[1]{Table~\ref{#1}}

\newcommand{\argmax}{\operatornamewithlimits{argmax}}
\newcommand{\argmin}{\operatornamewithlimits{argmin}}

\newcommand{\eg}{e.g. }
\newcommand{\ie}{i.e. }
\newcommand{\specialcell}[2][c]{%
  \begin{tabular}[#1]{@{}c@{}}#2\end{tabular}}

\title{Audio-Visual Speaker Diarization Based on Spatiotemporal Bayesian Fusion}
\author{Israel D. Gebru, Sil\`{e}ye Ba, Xiaofei Li and Radu Horaud
\IEEEcompsocitemizethanks{%I.~D. Gebru, S. Ba, X. Li and R. Horaud, 
INRIA Grenoble Rh\^one-Alpes, Montbonnot Saint-Martin, France. \protect\\
E-mail: \{israel-$ $-dejene.gebru, sileye.ba, xiaofei.li, radu.horaud\}@inria.fr}
\thanks{Funding from the European Union FP7 ERC Advanced Grant  VHIA (\#340113) and from 
XEROX University Affairs Committee (UAC) grant (2015-2017) is greatly acknowledged.}
}

\begin{document}
\maketitle
\begin{abstract}
Speaker diarization consists of assigning speech signals to people engaged in a dialogue.
An audio-visual spatiotemporal diarization model is proposed. The model is well suited for challenging scenarios that consist of several participants engaged in multi-party interaction while they move around and turn their heads towards the other participants rather than facing the cameras and the microphones. Multiple-person visual tracking  is combined with multiple speech-source localization in order to tackle the speech-to-person association problem. The latter is solved within a novel audio-visual fusion method on the following grounds: binaural spectral features are first extracted from a microphone pair, then a supervised audio-visual alignment technique maps these features onto an image, and finally a semi-supervised clustering method assigns binaural spectral features to visible persons. The main advantage of this method over previous work is that it processes in a principled way speech signals uttered simultaneously by multiple persons. 
The diarization itself is cast into a latent-variable temporal graphical model that infers speaker identities and speech turns, based on the output of an audio-visual association process, executed at each time slice, and on the dynamics of the diarization variable itself. The proposed formulation yields an efficient exact inference procedure. A novel dataset, that contains audio-visual training data as well as a number of scenarios involving several participants engaged in formal and informal dialogue, is introduced. The proposed method is thoroughly tested and benchmarked with respect to several state-of-the art diarization algorithms.

\end{abstract}
\section{Introduction}
\label{sec:introduction}
\input{introduction_v1}

\section{Related Work}
\label{sec:related}
\input{related_v1}

\section{Proposed Model}
\label{sec:model}
\input{model_v1}

\section{Visual Observations}
\label{sec:visual}
\input{visual_v1}

\section{Audio Observations}
\label{sec:audio}
\input{audio_v1}

\section{Audio-Visual Fusion}
\label{sec:audio-visual}
\input{audiovisual_v1}
\section{Audio-Visual Datasets}
\label{sec:audio-visual-data}
\input{audiovisualdata_v1}

\section{Experimental Evaluation}
\label{sec:experiments}
\input{experiments_v1}

\section{Conclusions}
\label{sec:conclusions}
We proposed an audio-visual diarization method well suited for challenging scenarios consisting of participants that either interrupt each other, or speak simultaneously. In both cases, the speech-to-person association problem is a difficult one. We proposed to combine multiple-person visual tracking with multiple speech-source localization in a principled spatiotemporal Bayesian fusion model. Indeed, the diarization process was cast into a latent-variable dynamic graphical model. We described in detail the derivation of the proposed model and we showed that, in the presence of a limited number of speakers (of the order of ten), the diarization formulation is efficiently solved via an exact inference procedure. Then we described a novel multiple speech-source localization method and a weakly supervised audio-visual clustering method. 

We also introduced a novel dataset, \textbf{AVDIAR}, that was carefully annotated and that enables to assess the performance of audio-visual (or audio-only) diarization methods using scenarios that were not available with existing datasets, \eg the participants were allowed to freely move in a room and to turn their heads towards the other participants, rather than always facing the camera. We also benchmarked our method with several other recent methods using publicly available datasets. Unfortunately, we were not able to compare our method with the methods of \cite{garau2010audio,noulas2012multimodal} for two reasons: first, these methods require long speech segments (of the order of 10~s), and second the associated software packages are not publicly available, which would have facilitated the comparison task. 

In the future we plan to incorporate richer visual features, such as head pose estimation and head-pose tracking, in order to facilitate the detection of speech turns on the basis of gaze or of people that look at each other over time.
We also plan to incorporate richer audio features, such as the possibility to extract speech signals emitted by each participant (sound-source separation) followed by speech recognition, and hence to enable not only diarization but also speech-content understanding.
\addnote[camera-network]{1}{Another extension is to consider distributed sensors, wearable devices, or a combination of both, in order to be able to deal with more complex scenarios involving tens of participants \cite{yan2013no,alameda2015analyzing}}.

\bibliographystyle{IEEEtran}
%\bibliography{IEEEabrv,mainref}
% Generated by IEEEtran.bst, version: 1.13 (2008/09/30)

\end{document}

%% file: introduction_v1.tex
% introduction-radu
In human-computer interaction (HCI) and human-robot interaction (HRI) it is often necessary to solve multi-party dialogue problems. For example, if two or more persons are engaged in a conversation, one important task to be solved, prior to automatic speech recognition (ASR) and natural language processing (NLP),  is to correctly assign temporal segments of speech to corresponding speakers. 
In the speech and language processing literature
this problem is referred to as \textit{speaker diarization}, or ``\textit{who speaks when}?"  A number of diarization methods were recently proposed, \eg \cite{anguera2012speaker}. 
If only unimodal data are available, the task is extremely difficult. Acoustic data are inherently ambiguous because they contain mixed speech signals emitted by several persons, corrupted by reverberations, by other sound sources and by background noise. Likewise, the detection of speakers from visual data is very challenging and it is limited to lip and facial motion detection from frontal close-range images of people: in more general settings, such as informal gatherings, people are not always facing the cameras, hence lip reading cannot be readily achieved. 

Therefore, an interesting and promising alternative consists of combining the merits of audio and visual data. The two modalities provide complementary information and hence audio-visual approaches to speaker diarization are likely to be more robust than audio-only or vision-only approaches. Several audio-visual diarization  methods have been investigated for the last decade, \eg\cite{garau2010audio,noulas2012multimodal,el2014audiovisual,minotto2015multimodal,sarafianos2016audio,kapsouras2016multimodal}. Diarization is based on audio-visual association, on the premise that a speech signal \textit{coincides} with the visible face of a speaker. This coincidence must occur both in space and time. 

In formal scenarios, \eg meetings, diarization is facilitated by the fact that participants take speech turns, which results in (i)~a clear-cut distinction between speech and non-speech and (ii)~the presence of short silent intervals between speech segments. Moreover, participants are seated, or are static, and there are often dedicated close-field microphones and cameras for each participant \eg\cite{carletta2005ami}. In these cases, the task consists of associating audio signals that contain clean speech with frontal images of faces: 
audio-visual association methods based on \textit{temporal coincidence} between the audio and visual streams seem to provide satisfactory results, \eg canonical correlation analysis  (CCA) \cite{kidron2005pixels,kidron2007cross,sargin2007audiovisual} or mutual information (MI) \cite{hershey2000audio,fisher2000learning,garau2010audio,noulas2012multimodal}. Nevertheless, temporal association between the two modalities is only effective on the premises that (i)~speech segments are uttered by a single person at a time, that (ii)~single-speaker segments are relatively long, and that (iii)~speakers continuously face the cameras. 

In informal scenarios, \eg ad-hoc social events, the audio signals are provided by distant microphones, hence the signals are corrupted by environmental noise and by reverberations. Speakers interrupt each other, hence short speech signals may occasionally be uttered simultaneously by different speakers. Moreover, people often wander around, turn their head away from the cameras, may be occluded by other people, suddenly appear or disappear  from the cameras'  fields of view, etc. Some of these problems were addressed in the framework of audio-visual speaker tracking, \eg \cite{gatica2007audiovisual,NavqiMiaoChambers2010,kilic2015audio}. Nevertheless, audio-visual tracking is mainly concerned with finding speaker locations and speaker trajectories, rather than solving the speaker diarization problem. 

In this paper it is proposed a novel spatiotemporal diarization model that is well suited for challenging scenarios that consist of several participants engaged in multi-party dialogue. The participants are allowed to move around and to turn their heads towards the other participants rather than facing the cameras. We propose to combine multiple-person visual tracking with multiple speech source localization in order to tackle the speech to person association problem. The latter is solved within a novel audio-visual fusion method on the following grounds: acoustic spectral features are extracted from a microphone pair, a novel supervised audio-visual alignment technique maps these features onto the image plane such that the audio and visual modalities are represented in the same mathematical space, a semi-supervised clustering method assigns the acoustic features to visible persons. The main advantage of this method over previous work is twofold: it processes in a principled way speech signals uttered simultaneously by multiple persons, and it enforces spatial coincidence between audio and visual features. 

Moreover, we cast the diarization process into a latent-variable temporal graphical model that infers over time both speaker identities and speech turns. This inference is based on combining the output of the proposed audio-visual fusion, that occurs at each time-step, with a dynamic model of the diarization variable (from the previous time-step to the current time-step), \ie a state transition model. We describe in detail the proposed formulation which is efficiently solved via an exact inference procedure. We introduce a novel dataset that contains audio-visual training data as well as a number of scenarios involving several participants engaged in formal and informal dialogue. We thoroughly test and benchmark the proposed method with respect to several state-of-the art diarization algorithms.

The remainder of this paper is organized as follows. Section~\ref{sec:related} describes the related work. Section~\ref{sec:model} describes in detail the temporal graphical model. Section \ref{sec:visual} describes visual feature detection and Section~\ref{sec:audio} describes the proposed audio features and their detection. Section~\ref{sec:audio-visual} describes the proposed semi-supervised audio-visual association method. The novel audio-visual dataset is presented in detail in Section~\ref{sec:audio-visual-data} while numerous experiments, tests, and benchmarks are presented in Section~\ref{sec:experiments}. Finally, Section~\ref{sec:conclusions} draws some conclusions. Videos, Matlab code and additional examples are available online.\footnote{\url{https://team.inria.fr/perception/avdiarization/}}

%% file: related_v1.tex
The task of speaker diarization is to detect speech segments and to group segments that correspond to the same speaker without any prior knowledge about the speakers involved nor their number. This can be done using auditory features alone, or a combination of auditory and visual features. Mel frequency cepstral coefficients (MFCC) is often the representation of choice whenever audio signal segments correspond to a single speaker. Then the diarization pipeline consists of splitting the audio frames into speech and non-speech frames, of extracting an MFCC feature vector from each speech frame and of performing agglomerative clustering such that each cluster found at the end corresponds to a different speaker \cite{wooters2008icsi}. Consecutive speech frames are assigned either to the same speaker and grouped into segments, or to different speakers, by using a state transition model, \eg HMM. 

\addnote[av-synchrony-1]{1}{The use of visual features for diarization has been motivated by the importance of audio-visual synchrony. Indeed, it was shown that facial and lip movements are strongly correlated with speech production \cite{yehia1998quantitative} and hence visual features, extracted from frontal views of speaker faces, can be used to increase the discriminative power of audio features in numerous tasks, \eg speech recognition \cite{potamianos2003recent}, source separation, \cite{rivet2007mixing,barzelay2010onsets} and diarization \cite{fisher2000learning,nock2003speaker,siracusa2007dynamic,noulas2007line}. In the latter case, the most common approaches involve the analysis of temporal correlation between the two modalities such that the face/lip movements that best correlate with speech correspond to an active speaker.}

\addnote[av-synchrony-2]{1}{Garau et al. \cite{garau2010audio} compare two audio-visual synchronization methods, based on mutual information (MI) and on canonical correlation analysis (CCA), and using MFCC auditory features combined with motion amplitude computed from facial feature tracks. They conclude that MI performs slightly better than CCA and that vertical facial displacements (lip and chin movements) are the visual features the most correlated with speech production. MI that combines gray-scale pixel-value variations extracted from a face region with acoustic energy is also used by Noulas et al. \cite{noulas2012multimodal}. The audio-visual features thus extracted are plugged into a dynamic Bayesian network (DBN) that perform speaker diarization. The method was tested on video meetings involving up to four participants which are recorded with several cameras, such that each camera faces a participant. More recently, both El Khoury et al. \cite{el2014audiovisual} and Kapsouras et al. \cite{kapsouras2016multimodal} propose to cluster audio features and face features independently and then to correlated these features based on temporal alignments between speech and face segments.} 

\addnote[clean-dirty]{1}{The methods mentioned so far yield good results whenever clean speech signals and frontal views of faces are available. A speech signal is said to be \textit{clean} if it is noise free and if it corresponds to a single speaker;  hence audio clustering based on MFCC (mel-frequency cepstral coefficients) features performs well. Moreover, time series of MFCC features seem to correlate well with facial-feature trajectories. If several faces are present, it is possible to select the facial feature trajectory that correlate the most with the speech signal, e.g. \cite{kidron2005pixels,kidron2007cross}. However, in realistic settings, participants are not always facing the camera, consequently the detection of facial and lip movements is problematic. Moreover, methods based on cross-modal temporal correlation, e.g. \cite{fisher2000learning,nock2003speaker,potamianos2003recent,siracusa2007dynamic,noulas2007line,noulas2012multimodal} require long sequences of audiovisual data, hence they can only be used offline such as the analysis of broadcast news, of audiovisual conferences, etc.}

%\addnote[ch:israel]{1}{Barzelay et al. \cite{barzelay2010onsets} who extend the earlier work of Kidron et al. \cite{kidron2005pixels} seek correspondence between features in the audio and video streams that exhibits strong temporal variation. Based on these features, they proposed a sequential matching procedure to pinpoint individual audio-associated visual objects pixel locations and isolate its corresponding audio component. It was used for speaker diarization in \cite{noulas2010audiovisual_phdthesis} and compared with that of \cite{noulas2012multimodal} on meeting videos.  Most recently, both El Khoury et al. \cite{el2014audiovisual} and Kapsouras et al. \cite{kapsouras2016multimodal} propose to cluster audio features and face features independently, prior to audio-visual correlation, which is based on the temporal alignment between speech segments and face segments. They showed that audio-visual diarization could be performed without the need for fine-grained pixel-based visual features.}

In the presence of simultaneous speakers, the task of diarization is more challenging because multiple-speaker information must be extracted from the audio data, one one hand, and the speech-to-face association problem must be properly addressed, on the other hand. In mixed-speech microphone signals, or dirty speech, there are many audio frames that contain acoustic features uttered by several speakers and MFCC features are not reliable anymore because they are designed to characterize acoustic signals uttered by single speakers. The multi-speech-to-multi-face association problem cannot be solved neither by performing temporal correlation between a single microphone signal and an image sequence nor by clustering MFCC features.

\addnote[multiple-speech]{1}{
One way to overcome the problems just mentioned is to perform multiple speech-source localization \cite{mandel2010,blandin2012multi,dorfan2015tree} and to associate speech sources with persons. These methods, however, do not address the problems of aligning speech-source locations with visible persons and of tracking them over time. Moreover, they often use circular or linear microphone arrays, e.g. \textit{planar} microphone setups, hence they provide sound-source directions with one degree of freedom, e.g. azimuth, which may not be sufficient to achieve robust audio-visual association. Hence, some form of microphone-camera calibration is needed. Khalidov et al. propose to 
estimate the microphone locations into a camera-centered coordinate system \cite{khalidov:hal-00861482} and to use a binocular-binaural setup in order to jointly cluster visual and auditory feature via a conjugate mixture model \cite{khalidov:inria-00590267}. Minotto et al. \cite{minotto2015multimodal} learn an SVM classifier using labeled audio-visual features. This training is dependent on the acoustic properties of experimental setup. They combine voice activity detection with sound-source localization using a linear microphone array which provides horizontal (azimuth) speech directions. In terms of visual features, their method relies on lip movements, hence frontal speaker views are required. 
}

Multiple-speaker scenarios were thoroughly addressed in the framework of audio-visual tracking. Gatica-Perez et al. \cite{gatica2007audiovisual} proposed a multi-speaker tracker using approximate inference implemented with a Markov chain Monte Carlo particle filter (MCMC-PF). 
Navqi et al. \cite{NavqiMiaoChambers2010} proposed a 3D visual tracker, based as well on MCMC-PF, to estimate the positions and velocities of the participants which are then passed to blind source separation based on beamforming \cite{van1988beamforming}. 
%While this provides a proof-of-concept benchmark for moving speakers, MCMC-PF tracking cannot easily handle a varying number of speakers.
Reported experiments of both \cite{gatica2007audiovisual,NavqiMiaoChambers2010} require a network of distributed cameras to guarantee that frontal views of the speakers are always available. More recently, Kilic et al. \cite{kilic2015audio} proposed to use audio information to assist the particle propagation process and to weight the observation model. This implies that audio data are always available and that they are reliable enough to properly relocate the particles. While audio-visual multiple-person tracking methods provide an interesting methodology, they do not address the diarization problem. Indeed, they assume that people speak continuously, which facilitates the task of the proposed audio-visual trackers. With the exception of \cite{NavqiMiaoChambers2010}, audio analysis is reduced to sound-source localization using a microphone array, and this in order to enforce spatial coincidence between faces and speech.

Recently we addressed audio-visual speaker diarization under the assumption that participants take speech turns and that there is no overlap between their emitted speech signals. We proposed a simple model that consists of a speech-turn discrete latent variable that associates the speech signal with one of the participants \cite{gebru:hal-01163659,gebru:hal-01220956}. The main idea of this work was to track multiple persons and to extract a single sound-source direction from short time intervals, \eg using \cite{deleforge2015colocalization} to map sound directions onto the image plane. Audio and visual observations can then be associated using a recently proposed weighted-data EM algorithm \cite{gebru2016em}. In the present paper we propose a novel dynamic audio-visual fusion model that can deal with simultaneously speaking participants. In particular, we exploit the spectral sparsity of speech signals and we propose a novel multiple speech source localization method based on a semi-supervised complex-Gaussian mixture model in the Fourier domain. We also generalize the single speaker-turn diarization model of \cite{gebru:hal-01163659,gebru:hal-01220956} to multiple speaking persons.

Recently we addressed audio-visual speaker diarization under the assumption that participants take speech turns and that there is no overlap between their speech segments. We proposed a model that consists of a speech-turn discrete latent variable that associates the current speech signal, if any, with one of the visible participants \cite{gebru:hal-01163659,gebru:hal-01220956}. The main idea was to perform multiple-person tracking in the visual domain, to extract sound-source directions (one direction at a time), and to map this sound direction onto the image plane \cite{deleforge2015colocalization}. Audio and visual observations can then be associated using a recently proposed weighted-data EM algorithm \cite{gebru2016em}. 

\addnote[rel:originality]{1}{
In this present paper we propose a novel DBN-based cross-modal diarization model. Unlike several recently proposed audio-visual  diarization works \cite{noulas2012multimodal}, \cite{el2014audiovisual}, \cite{kapsouras2016multimodal}, \cite{gebru:hal-01163659,gebru:hal-01220956}, the proposed model 
can deal with simultaneously speaking participants that may wander around and turn their faces away from the cameras. Unlike \cite{noulas2012multimodal}, \cite{el2014audiovisual}, \cite{kapsouras2016multimodal} which require long sequences of past, present, and future frames, and hence are well suited for post-processing, our method is \textit{causal} and therefore it can be used online.  To deal with mixed speech signals, we exploit the sparsity of speech spectra and we propose a novel multiple speech-source localization method based on audio-visual data association implemented  with a cohort of frequency-wise semi-supervised complex-Gaussian mixture models.
}

%% file: model_v1.tex
We start by introducing a few notations and definitions. Unless otherwise specified, upper-case letters denote random variables while lower-case letters denote their realizations. Vectors are in slanted bold, \eg $\Xvect, \Yvect$, while matrices are in bold, \eg $\Xmat, \Ymat$.
We consider an image sequence that is synchronized with two microphone signals and let $t$ denote the time-step index of the audio-visual stream of data. 
% the term time-step is more common than time-slice

Let $N$ be the maximum number of visual objects, \eg persons, available at any time $t$. Hence at $t$ we have at most $N$ persons with locations on the image plane $\Xmat_t = (\Xvect_{t,1}, \dots, \Xvect_{t,n}, \dots, \Xvect_{t,N})\in\mathbb{R}^{2\times N}$, where the observed random variable $\Xvect_{t,n}\in\mathbb{R}^{2}$ is the pixel location of person $n$ at $t$. We also introduce a set of binary (or control) variables $\Vvect_t = (V_{t,1}, \dots V_{t,n}, \dots V_{t,N})\in \{0,1\}^{N} $ such that $V_{t,n}=1$ if person $n$ is \textit{visible} at $t$ and $V_{t,n}=0$ if the person is not visible. Let $N_t=\sum_n V_{t,n}$ denote the number of visible persons at $t$. The time series $\Xmat_{1:t} = \{ \Xmat_1, \dots, \Xmat_t\}$ and associated \textit{visibility binary masks} $\Vmat_{1:t}=\{\Vvect_1, \dots, \Vvect_t\}$ can be estimated using a multi-person tracker, \ie Section~\ref{sec:visual}. %We perform multi-person tracking using \cite{bae2014robust} (see section~\ref{sec:implementation} below).

We now describe the audio data. Without loss of generality, the audio signals are recorded with two microphones: let $\Ymat_t=(\Yvect_{t,1}, \dots, \Yvect_{t,k}, \dots, \Yvect_{t,K})\in\mathbb{C}^{F\times K}$ be a \textit{binaural spectrogram} containing $F$ number of frequencies and $K$ number of frames. Each frame is a binaural vector $\Yvect_{t,k}\in\mathbb{C}^{F}, 1\leq k \leq K$.  Binaural spectrograms are obtained in the following way. The short-time Fourier transform (STFT) is first applied to the left- and right-microphone signals acquired at time-step $t$ such that two spectrograms, $\Lmat_t, \Rmat_t \in\mathbb{C}^{F\times K}$ are associated with the left and right microphones, respectively. Each spectrogram is composed of $F\times K$ complex-valued STFT coefficients. The binaural spectrograms $\Ymat_t$ is composed of $F\times K$ complex-valued coefficients and each coefficients $Y_{t,k}^{f}, 1\leq f \leq F$ and $1\leq k \leq K$, can be estimated from the corresponding left- and right-microphone STFT coefficients $L_{t,k}^{f}$ and $R_{t,k}^{f}$, \ie Section~\ref{sec:audio}. One important characteristic of speech signals is that they have sparse spectrograms. As explained below, this sparsity is explicitly exploited by the proposed speech-source localization method. Moreover, the microphone signals are obviously contaminated by background noise and by sounds emitted by other non-speech sources. Therefore, \textit{speech activity} associated with each binaural spectrogram entry $Y_{t,k}^{f}$ must be properly detected and characterized with the help of a binary-mask matrix $\Amat_t\in \{0,1\}^{F\times K}$: $A_{t,k}^{f} =1$ if the corresponding spectrogram coefficient contains speech, and $A_{t,k}^{f} =0$ if it does not contain speech. To summarize, the binaural spectrograms $\Ymat_{1:t} = \{ \Ymat_1, \dots, \Ymat_t\}$ and associated \textit{speech-activity masks} $\Amat_{1:t} = \{ \Amat_1, \dots, \Amat_t\}$ characterize the audio observations. 

\input{model_tikz}
\subsection{Speaker Diarization Model}
\label{sec:tracking-multiple}

We remind that the objective of our work is to assign speech signal to persons, which amounts to one-to-one spatiotemporal associations between several speech sources (if any) and one or several observed persons. For this purpose we introduce a time series of discrete latent variables, $\Smat_{1:t}=\{\Svect_1, \dots, \Svect_t\}\in \{0,1\}^{N\times t}$ where the vector $\Svect_t= (S_{t,1}, \dots, S_{t,n}, \dots, S_{t,N})\in \{0,1\}^{N}$ has binary-valued entries such that $S_{t,n}=1$ if person $n$ \textit{speaks} during the time-step  $t$, and $S_{t,n}=0$ if person $n$ is \textit{silent}. \addnote[ch:israel]{1}{The temporal speaker diarization problem at hand can be formulated as finding a maximum-a-posteriori (MAP) solution, namely finding the most probable configuration of the latent state $\Svect_t$ that maximizes the following posterior probability distribution, also referred to as the filtering distribution:}
\begin{equation}
\label{eq:map-speaker}
\hat{\svect}_t = \argmax_{\svect_t} P (\Svect_t = \svect_t| \xmat_{1:t}, \ymat_{1:t}, \vmat_{1:t}, \amat_{1:t}).
\end{equation}
We introduce the notation $\Umat_t = (\Xmat_t, \Ymat_t, \Amat_t)$ for the observed variables, while the $\Vmat_t$ are referred to as control variables.
The filtering distribution \eqref{eq:map-speaker} can be expanded as:
\begin{align}
P (\smat_t | \umat_{1:t}, \vmat_{1:t})  &= 
 \frac{P(\umat_t | \smat_t, \umat_{1: t-1},\vmat_{1:t}) P(\smat_t | \umat_{1: t-1}, \vmat_{1:t})}
{P(\umat_{t} | \umat_{1:t-1}, \vmat_{1:t})} \nonumber \\
\label{eq:MAP-dev1}
&= 
 \frac{P(\umat_t | \smat_t,\vvect_{t}) P(\smat_t | \umat_{1: t-1}, \vmat_{1:t})}
{P(\umat_{t} | \umat_{1:t-1}, \vmat_{1:t})}.
\end{align}
\addnote[ch:israel]{1}{We assumed that the observed variables $\Umat_t$ are conditionally independent of all other variables, given the speaking state $\Svect_t$ and control input $\Vmat_t$; $\Svect_t$ is conditionally independent of $\Svect_1,\ldots,\Svect_{t-2}$, given $\Svect_{t-1}$ and $\Vvect_{t-1:t}$. Fig.~\ref{fig:graphicalModel} shows the graphical model representation of the proposed model.}

The numerator of \eqref{eq:MAP-dev1} is the product of two terms: the observation likelihood (left) and the predictive distribution (right). The observation likelihood can be expanded as:
\begin{align}
P(\umat_t | \smat_t,\vvect_{t}) = \prod_{n=1}^{N} \bigg( & P(\umat_t | S_{t,n}=1, V_{t,n})^{s_{t,n}} 
\nonumber \\
\label{eq:likelihood-expanded}
& \times P(\umat_t | S_{t,n}=0,V_{t,n})^{1- s_{t,n}} \bigg).
\end{align}
The predictive distribution (right hand side of the numerator of \eqref{eq:MAP-dev1}) expands as:
\begin{align}
P(\smat_t | & \umat_{1: t-1}, \vmat_{1:t}) \nonumber \\
=& \sum_{\smat_{t-1}} P(\smat_t , \smat_{t-1} | \umat_{1: t-1}, \vmat_{1:t}) 
\nonumber \\
=& \sum_{\smat_{t-1}} P(\smat_t | \smat_{t-1}, \umat_{1: t-1}, \vmat_{1:t}) P(\smat_{t-1} | \umat_{1: t-1}, \vmat_{1:t}) 
\nonumber \\
=& \sum_{\smat_{t-1}} P(\smat_t | \smat_{t-1}, \vvect_{t}, \vvect_{t-1}) P(\smat_{t-1} | \umat_{1: t-1}, \vmat_{1:t-1}) 
\nonumber \\
\label{eq:state-dynamics}
=& \sum_{\smat_{t-1}} \bigg( \prod_{m=1}^{N} P( s_{t,m} | s_{t-1,m}, v_{t,m}, v_{t-1,m}) \bigg) \\
\label{eq:map-previous}
& \times P(\smat_{t-1} | \umat_{1: t-1}, \vmat_{1:t-1}),
\end{align}
which is the product of the state transition probabilities \eqref{eq:state-dynamics} and of the filtering distribution at $t-1$ \eqref{eq:map-previous}.
We now expand the denominator of \eqref{eq:MAP-dev1}:
\begin{align}
\label{eq:denominator}
P(\umat_{t} | \umat_{1:t-1}, \vmat_{1:t}) &= \sum_{\smat_{t}} P(\umat_{t},\smat_{t} | \umat_{1:t-1}, \vmat_{1:t}) \nonumber \\
&=\sum_{\smat_{t}} P(\umat_t | \smat_t,\vmat_{t}) P(\smat_{t} | \umat_{1: t-1}, \vmat_{1:t}).
\end{align}
\addnote[complexity]{1}{To summarize, the evaluation of the filtering distribution at an arbitrary time-step $t$ requires the evaluation of (i)~the observation likelihood \eqref{eq:likelihood-expanded}, \ie Section~\ref{sec:audio-visual}, (ii)~the state transition probabilities \eqref{eq:state-dynamics}, \ie Section~\ref{sec:state-transition}, (iii)~ the filtering distribution at $t-1$ \eqref{eq:map-previous}, and of (iv) the normalization term \eqref{eq:denominator}. Notice that the number of possible state configuration is $2^N$ where $N$ is the maximum number of people. For small values of $N$ (2 to 6 persons), solving the MAP problem \eqref{eq:map-speaker} is computationally efficient}.

\subsection{State Transition Model}
\label{sec:state-transition}
Priors over the dynamics of the state variables in  \eqref{eq:state-dynamics} exploit the simplifying assumption that the speaking dynamics of a person is independent of all the other persons. \addnote[turn-taking]{1}{Several existing speech-turn models rely on non-verbal cues, such as filled pauses, breath, facial gestures, gaze, etc. \cite{bohus2011decisions,skantze2014turn}, and a speech-turn classifier can be built from annotated dialogues. The state transition model of \cite{noulas2012multimodal} considers all possible transitions, e.g., speaking/non-speaking, visible/not-visible, etc., which results in a large number of parameters that need be estimated. These models cannot be easily extended when there are speech overlaps and one has to rely on features extracted from the data}. 
%Although in a linguistic perspective this is not always the case; we do take into account the state of the other speakers while speaking but from a data modeling point of view, it is a practical assumption taken to reduce the number of parameters needed to describe the state transition model.
To define the speaking transition priors $P( s_{t,n} | s_{t-1,n}, v_{t,n}, v_{t-1,n})$, we consider three cases: (i)~person $n$ visible at $t-1$ and visible at $t$, or $v_{t,n}= v_{t-1,n}=1$ and in this case the transitions are parametrized by a self-transition prior $q \in [0,1]$ which models the probability to remain in the same state, either speaking or not speaking, (ii)~person $n$ not visible at $t-1$ and visible at $t$, or $v_{t,n}=1, v_{t-1,n}=0$, in this case, the prior to be either speaking or not speaking at $t$ is uniform, and (iii)~person $n$ not visible at $t$, or  $v_{t,n}=0, v_{t-1,n}=1$, in which case the prior not to be speaking is equal to 1. The following equation summarizes all these cases:
\begin{align}
\label{eq:transition-cases}
P( s_{t,n} | & s_{t-1,n}, v_{t,n}, v_{t-1,n})  \nonumber \\
&= v_{t,n}v_{t-1,n}q^{\delta_{s_{t-1,n}}(s_{t,n})}(1-q)^{1-\delta_{s_{t-1,n}}(s_{t,n})} \nonumber \\
&+ \frac{1}{2}(1-v_{t-1,n})v_{t,n} + (1-v_{t,n})\delta_0(s_{t,n}),
\end{align}
where $\delta_i (j) = 1$ if $i=j$ and $\delta_i (j) = 0$ if $i\neq j$. \addnote[state-transition-1]{1}{Note that this does not consider the case of person $n$ not visible at $t-1$ and at $t$ for which the prior probability to be speaking is 0. In all our experiments we used $q=0.8$}.

\addnote[state-transition-2]{1}{The multiple-speaker tracking and diarization model proposed in this work only considers persons that are both seen and heard. Indeed, in informal scenarios there may be acoustic sources (speech or other sounds such as music) that are neither in the camera field of view, nor can they be visually detected and tracked. The proposed audio-visual association model addresses this problem, i.e. Section~\ref{sec:audio-visual}.}

%According to our model, non-visible persons are considered as if they are not speaking, thus diarization is performed on people that are both seen and heard. The estimation of the visibility variables by the visual tracker is not always perfect, especially in challenging scenarios where occlusions often lead to tracking failures. When the visibility variable for a speaking person is inaccurate, this person gets a zero posterior probability to be in a speaking state and this in turn leads to diarization miss error. Hence, a robust multi-target visual tracker is absolutely necessary. Moreover, in real world settings there are all other kinds of speech sources or general sound sources which are not in fact visually tracked, so without loss of generality they are said to be not-visible. The most challenging task is how to deal with those kind of not-visible (thus undesirable) sound sources since the recorded audio signal contains a mixture of desirable and undesirable sound sources. We solved this challenge by designing an observation model that discard in a principled way auditory observations that might be generated by non-visible sources (including non-visible but speaking persons), \ie Section~\ref{sec:audio-visual}.
%The observation likelihood of a non-visibility person conditioned on speaking state will get a zero probability.

%% file: model_tikz.tex
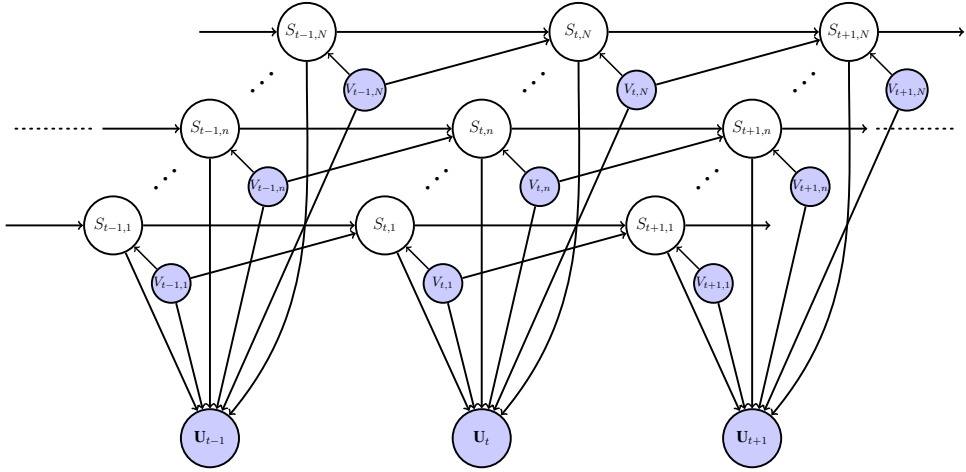
\begin{figure*}[Htp!]
\centering
\resizebox{1.5\columnwidth}{!}{%
\begin{tikzpicture}
[remember picture,
  hiddenBig/.style={circle,draw=black!200,line width=0.5mm,inner sep=0pt,minimum size=1.5cm},
  hiddenBigT/.style={circle,draw=black!200,line width=0.5mm},
  hiddenSmall/.style={circle,draw=black!200,line width=0.5mm},
  hiddenSmallT/.style={circle,draw=black!200,line width=0.5mm},
  Observation/.style={circle,draw=black!200,fill=blue!20,line width=0.5mm,minimum size=1.5cm},
  ObservationSmall/.style={circle,draw=black!200,fill=blue!20,line width=0.5mm,inner sep=0pt,minimum size=1cm},
  ]
\node (tm1) {
\begin{tikzpicture}
\node [hiddenBig,draw=black] (stm13)  at (2.5,2.5) [fontscale=1] {$S_{t-1,N}$};
\node [ObservationSmall,draw=black] (vtm13) at (4.0,1.0) {$V_{t-1,N}$};
\draw[black,line width=0.4mm,->] (vtm13)-- (stm13);  

\node [hiddenBig,draw=black] (stm12)  at (0,0) [fontscale=1] {$S_{t-1,n}$};
\node [ObservationSmall,draw=black] (vtm12) at (1.5,-1.5) {$V_{t-1,n}$};
\draw[black,line width=0.4mm,->] (vtm12)-- (stm12);

\node [hiddenBig,draw=black] (stm11)  at (-2.5,-2.5) [fontscale=1] {$S_{t-1,1}$};
\node [ObservationSmall,draw=black] (vtm11) at (-1.0,-4.0) {$V_{t-1,1}$};
\draw[black,line width=0.4mm,->] (vtm11)-- (stm11); 
 
\node [Observation,draw=black] (utm1)  at (0.0,-8) [fontscale=1] {$\Umat_{t-1}$};
\draw[black,line width=0.5mm,->] (stm13) to[out=270,in=50] (utm1); 
\draw[black,line width=0.5mm,->] (stm12) -- (utm1); 
\draw[black,line width=0.5mm,->] (stm11) -- (utm1); 
\draw[black,line width=0.5mm,->] (vtm13) -- (utm1); 
\draw[black,line width=0.5mm,->] (vtm12) -- (utm1); 
\draw[black,line width=0.5mm,->] (vtm11) -- (utm1); 
\end{tikzpicture}
 };
 \node [xshift=70mm] (t) {
\begin{tikzpicture}
\node [hiddenBigT,draw=black,minimum size=1.5cm] (st3)  at (2.5,2.5) [fontscale=1] {$S_{t,N}$};
\node [ObservationSmall,draw=black] (vt3) at (4.0,1.0) {$V_{t,N}$};
\draw[black,line width=0.4mm,->] (vt3)-- (st3);  

\node [hiddenBigT,draw=black,minimum size=1.5cm] (st2)  at (0,0) [fontscale=1] {$S_{t,n}$};
\node [ObservationSmall,draw=black,minimum size=1cm] (vt2) at (1.5,-1.5) {$V_{t,n}$};
\draw[black,line width=0.4mm,->] (vt2)-- (st2);

\node [hiddenBigT,draw=black,minimum size=1.5cm] (st1)  at (-2.5,-2.5) [fontscale=1] {$S_{t,1}$};
\node [ObservationSmall,draw=black] (vt1) at (-1.0,-4.0) {$V_{t,1}$};
\draw[black,line width=0.4mm,->] (vt1)-- (st1); 
 
\node [Observation,draw=black] (ut)  at (0.0,-8) [fontscale=1] {$\Umat_{t}$};
\draw[black,line width=0.5mm,->] (st3) to[out=270,in=50] (ut); 
\draw[black,line width=0.5mm,->] (st2)-- (ut); 
\draw[black,line width=0.5mm,->] (st1)-- (ut); 
\draw[black,line width=0.5mm,->] (vt3)-- (ut); 
\draw[black,line width=0.5mm,->] (vt2)-- (ut); 
\draw[black,line width=0.5mm,->] (vt1)-- (ut); 
\end{tikzpicture}
 };
 
  \node [xshift=140mm] (tp1) {
\begin{tikzpicture}
\node [hiddenBig,draw=black] (stp13)  at (2.5,2.5) [fontscale=1] {$S_{t+1,N}$};
\node [ObservationSmall,draw=black] (vtp13) at (4.0,1.0) {$V_{t+1,N}$};
\draw[black,line width=0.4mm,->] (vtp13)-- (stp13);  

\node [hiddenBig,draw=black] (stp12)  at (0,0) [fontscale=1] {$S_{t+1,n}$};
\node [ObservationSmall,draw=black] (vtp12) at (1.5,-1.5) {$V_{t+1,n}$};
\draw[black,line width=0.4mm,->] (vtp12)-- (stp12);

\node [hiddenBig,draw=black] (stp11)  at (-2.5,-2.5) [fontscale=1] {$S_{t+1,1}$};
\node [ObservationSmall,draw=black] (vtp11) at (-1.0,-4.0) {$V_{t+1,1}$};
\draw[black,line width=0.4mm,->] (vtp11)-- (stp11); 
 
\node [Observation,draw=black] (utp1)  at (0.0,-8) [fontscale=1] {$\Umat_{t+1}$};
\draw[black,line width=0.5mm,->] (stp13) to[out=270,in=50] (utp1); 
\draw[black,line width=0.5mm,->] (stp12)-- (utp1); 
\draw[black,line width=0.5mm,->] (stp11)-- (utp1); 
\draw[black,line width=0.5mm,->] (vtp13)-- (utp1); 
\draw[black,line width=0.5mm,->] (vtp12)-- (utp1); 
\draw[black,line width=0.5mm,->] (vtp11)-- (utp1); 
\end{tikzpicture}
 };

\draw[black,line width=0.5mm,->] (stm13)-- (st3); 
\draw[black,line width=0.5mm,->] (stm12)-- (st2); 
\draw[black,line width=0.5mm,->] (stm11)-- (st1); 
 
\draw[black,line width=0.5mm,->] (st3)-- (stp13); 
\draw[black,line width=0.5mm,->] (st2)-- (stp12); 
\draw[black,line width=0.5mm,->] (st1)-- (stp11); 

% this draw the arrow and  dots at t-1 and t+1
\node (A) [left=2cm of stm13] {};
\node (B) [left=2cm of stm12] {};
\node (C) [left=2cm of stm11] {};
\draw[black,line width=0.5mm,->] (A) -- (stm13); 
\draw[black,line width=0.5mm,->] (B) -- (stm12);
\draw[black,line width=0.5mm,->] (C) -- (stm11);

\node (D) [right=2.2cm of stp13] {};
\node (E) [right=2.2cm of stp12] {};
\node (F) [right=2.2cm of stp11] {};
\draw[black,line width=0.5mm,->] (stp13) -- (D); 
\draw[black,line width=0.5mm,->] (stp12) -- (E);
\draw[black,line width=0.5mm,->] (stp11) -- (F);

\node (G) [left=2cm of B] {};
\node (H) [right=2cm of E] {};
\draw[black,line width=0.5mm,loosely dotted,line cap=round] (G) -- (B);
\draw[black,line width=0.5mm,loosely dotted,line cap=round] (E) -- (H);

%% this draw the digonal links
\path (stm12) -- node[auto=false, fontscale=6]{\rotatebox[origin=c]{62}{$\ddots$}} (stm11);
\path (stm12) -- node[auto=false, fontscale=6]{\rotatebox[origin=c]{62}{$\ddots$}} (stm13);

\path (st2) -- node[auto=false, fontscale=6]{\rotatebox[origin=c]{62}{$\ddots$}} (st1);
\path (st2) -- node[auto=false, fontscale=6]{\rotatebox[origin=c]{62}{$\ddots$}} (st3);

\path (stp12) -- node[auto=false, fontscale=6]{\rotatebox[origin=c]{62}{$\ddots$}} (stp11);
\path (stp12) -- node[auto=false, fontscale=6]{\rotatebox[origin=c]{62}{$\ddots$}} (stp13);

%% this draw vt-1 to st
\draw[black,line width=0.5mm,->] (vtm11) -- (st1); 
\draw[black,line width=0.5mm,->] (vtm12) -- (st2); 
\draw[black,line width=0.5mm,->] (vtm13) -- (st3); 

\draw[black,line width=0.5mm,->] (vt1) -- (stp11); 
\draw[black,line width=0.5mm,->] (vt2) -- (stp12); 
\draw[black,line width=0.5mm,->] (vt3) -- (stp13);

\node (I) [below right=0.4cm and 0.2cm of F] {};
\node (J) [below right=0.4cm and 0.2cm of E] {};
\node (K) [below right=0.4cm and 0.2cm of D] {};

% \draw[black,line width=0.5mm,->] (vtp11) -- (I); 
% \draw[black,line width=0.5mm,->] (vtp12) -- (J); 
% \draw[black,line width=0.5mm,->] (vtp13) -- (K);

\end{tikzpicture}
}%
\caption{The Bayesian spatiotemporal fusion model used for audio-visual speaker diarization. Shaded nodes represent the observed variables, while unshaded nodes represent latent variables. Note that the visibility-mask variables $V_{t,n}$ although observed, they are treated as control variables. This model enables simultaneously speaking persons, which is not only a realistic assumption but also very common in natural dialogues and applications like for example HRI.}
\label{fig:graphicalModel}
\end{figure*}

%% file: visual_v1.tex
%visual
We propose to use visual tracking of multiple persons in order to infer realizations of the random variables $\Xmat_{1:t}$ introduced above. The advantage of a multiple-person tracker is that it is able to detect a variable number of persons, possibly appearing and disappearing from the visual field of view,  to estimate their velocities, and to track their locations and identities. Multiple object/person tracking is an extremely well studied topic in the computer vision literature and many methods with their associated software packages are available. Among all these methods, we chose the multiple-person tracker of \cite{bae2014robust}. In the context of our work, this method has several advantages: (i)~it robustly handles fragmented tracks (due to occlusions, to the limited camera field of view, or simply to unreliable detections), (ii)~it handles changes in person appearance, such as a person that faces the camera and then suddenly turns his/her head away from the camera, \eg towards a speaker, and (iii)~it performs online discriminative learning such that it can distinguish between similar appearances of different persons. 

\addnote[visual-tracking]{1}{Visual tracking is implemented in the following way. Un upper-body detector \cite{bourdev2009poselets} is used to extract bounding boxes of persons in every frame. This allows the tracker to initialize new tracks, to re-initialize lost ones, to avoid tracking drift, and to cope with a large variety of poses and resolutions. Moreover, an appearance model, based on the color histogram of a bounding box associated with a person upper body (head and torso), is associated with each detected person. The appearance model is updated whenever the upper-body detector returns a reliable bounding box (no overlap with another bounding box).
We observed that upper-body detection is more robust than face detection which yields many false positives. Nevertheless, in the context of audio-visual fusion, the face locations are important. Therefore, the locations estimated by the tracker, $\Xmat_{1:t}$, correspond to the face centers of the tracked persons. }

%% file: audio_v1.tex
%audio observations
In this section we present a methodology for extracting binaural features in the presence of either a single audio source or several speech sources.
We consider audio signals recorded with a binaural microphone pair. As already explained in Section~\ref{sec:model}, the short-time Fourier transform (STFT) is applied to the two microphone signals acquired at time-slice $t$ and two spectrograms are thus obtained, namely $\Lmat_t, \Rmat_t \in\mathbb{C}^{F\times K}$. 

\subsection{Single Audio Source}
\label{sec:single-speaker}

Let's assume that there is a single (speech or non-speech) signal emitted by an audio source during the time slice $t$. In the STFT domain, the relationships between the source-STFT spectrogram and microphone-STFT spectrograms are, for each frame $k$ and each frequency $f$ (for convenience we omit the time index $t$):
\begin{align}
\label{eq:left-microphone-one}
L_{k}^{f} &= H_{L,k}^{f} T_{k}^{f} + N_{L,k}^{f}\\
\label{eq:right-microphone-one}
R_{k}^{f}&= H_{R,k}^{f} T_{k}^{f} + N_{R,k}^{f},
\end{align}
where $\Tmat = \{ T_{k}^{f}\}_{k=1,f=1}^{k=K,f=F}$ is the unknown source spectrogram, $\Nmat_L = \{ N_{L,k}^{f}\}_{k=1,f=1}^{k=K,f=F}$ and $\Nmat_R = \{ N_{R,k}^{f}\}_{k=1,f=1}^{k=K,f=F}$ are the unknown noise spectrograms associated with the left and right channels, and $\Hmat_L=\{ H_{L,k}^{f}\}_{k=1,f=1}^{k=K,f=F}$ and $\Hmat_R=\{ H_{R,k}^{f}\}_{k=1,f=1}^{k=K,f=F}$ are the unknown left and right \textit{acoustic transfer functions} that are frequency-dependent. The above equations correspond to the general case of a moving sound source.
However, if we assume that the audio source is static during the time slice $t$, \ie the source emitter is in a fixed position during the time slice $t$, the acoustic transfer functions are time-invariant and only depend on the source position relative to the microphones. We further define \textit{binaural features}, \ie the ratio between the left and right acoustic transfer functions, $H_{L}^{f}/H_{R}^{f}$. Notice that we omitted the frame index because in the case of a static source, the acoustic transfer function is invariant over frames. Likewise the acoustic transfer function, the binaural features do not depend on $k$ and they only contain audio-source position information \cite{deleforge2015colocalization}. 

One can use the estimated cross-PSD (power spectral density) and auto-PSD to extract binaural features in the following way. The cross-PSD between the two microphones is \cite{li:hal-01119186,li:hal-01163675}:
\begin{align}
\label{eq:cross-psd}
\Phi_{L,R}^{f} & = \frac{1}{K} \sum_{k=1}^{K}L_{k}^{f}R_{k}^{f\star} \\
\label{eq:cross-psd-one}
& \approx \frac{1}{K} H_{L}^{f} H_{R}^{f\star} \sum_{k=1}^{K} |T_{k}^{f} |^2 + \frac{1}{K}  \sum_{k=1}^{K} N_{L,k}^{f} N_{R,k}^{f\ast},
\end{align}
where $A^\star$ is the complex-conjugate of $A$ and it is assumed that the signal-noise cross terms can be neglected. If the noise signals are spatially uncorrelated then the noise-noise cross terms can also be neglected. The binaural feature vector at $t$ can be approximated with the ratio between the cross-PSD and auto-PSD functions, \ie the vector $\Yvect_t= (Y_{t}^{1}, \dots Y_{t}^{f}, \dots Y_{t}^{F})\tp$ with entries
\begin{align}
\label{eq:RTF-one}
Y_{t}^{f} = \frac{\Phi_{t,L,R}^{f}}{\Phi_{t,R,R}^{f}}
\end{align}

\subsection{Multiple Speech Sources}
\label{sec:multiple-speakers}

% Xiaofei: I suggest we start the signal formulation here with multiple speakers
We now consider the case of $P$ speakers ($P>1$) that emit speech signals simultaneously (for convenience we omit again the time index $t$)
\begin{align}
\label{eq:left-microphone-several}
L_{k}^{f} &= \sum_{p=1}^P H_{L,p}^{f} T_{p,k}^{f} + N_{L,k}^{f}\\
\label{eq:right-microphone-several}
R_{k}^{f}&= \sum_{p=1}^P H_{R,p}^{f} T_{p,k}^{f} + N_{R,k}^{f},
\end{align}
where $H_{L,p}^{f}$ and $H_{R,p}^{f} $ are the acoustic transfer functions from the speech-source $p$ to the left and right microphones, respectively. 
% Xiaofei: we have two options: 
% (1) Directly extract the rtf from each time-frequency bin Y_{t,k}^{f} = \frac{L_{k}^{f}}{R_{k}^{f}}. 
% (2) First compute the cross- and auto-PSD using the past D frames, then compute Y_{t,k}^{f} using eq:RTF-several. I think the speech sparsity will be preserved to a great extent.
%   The speech and non-speech frames classification presented in the last ICASSP paper can be used for reducing noise by spectral subtraction, and binary-masking in eq:binary-mask.
The STFT based estimate of the cross-PSD for each frequency-frame point $(f,k)$ is 
\begin{align}
\label{eq:cross-psd-several}
\Phi_{L,R,k}^{f}  = L_{k}^{f}R_{k}^{f\star}.
\end{align}
In order to further characterize simultaneously emitting speech signals, we exploit the well-known fact that speech signals have sparse spectrograms in the Fourier domain. Because of this sparsity it is realistic to assume that only one speech source $p$ is active at each frequency-frame point of the two microphone spectrograms \eqref{eq:left-microphone-several} and \eqref{eq:right-microphone-several}. Therefore these spectrograms are composed of STFT coefficients that contain (i)~either speech emitted by a single speaker, (ii)~or noise. Using this assumption, the binaural spectrogram $\Ymat_t$ and associated binary mask matrix $\Amat_t$ can be estimated from the cross-PSD and auto-PSD in the following way. We start by estimating a binary mask for each frequency-frame point, %\eg \cite{li:hal-01119186} ?????
\begin{align}
\label{eq:binary-mask}
A_{k}^{f} = 
\begin{cases}
0 & \text{if $\max (\Phi_{L,L,k}^{f}, \Phi_{R,R,k}^{f}) < a$} \\
1 & \text{otherwise},
\end{cases}
\end{align}
\addnote[adaptive-th]{1}{where $a$ is an adaptive threshold whose value is estimated based on noise statistics \cite{li2016non}. Then, we compute the binaural spectrogram coefficients for each frequency-frame point $(f,k)$ at time-slice $t$ as:}
\begin{align}
\label{eq:RTF-several}
%\begin{array}{ll}
Y_{t,k}^{f} = 
\begin{cases} \frac{\Phi_{t,L,R,k}^{f}}{\Phi_{t,R,R,k}^{f}} & \text{if} \quad A_{t,k}^{f} = 1 \\
0 & \text{if} \quad A_{t,k}^{f} = 0.
\end{cases}
%\end{array}
\end{align}
It is important to stress that while these binaural coefficients are source-independent, they are location-dependent. This is to say that the binaural spectrogram only contains information about the location of the sound source and not about the content of the source. This crucial property allows one to use different types of sound sources for training a sound source localizer and for predicting the location of a speech source, as explained in the next section.

%% file: audiovisual_v1.tex
% audiovisual
% israel: training dataset to training data?
% image location to pixel location/coordinate
In this section we propose an audio-visual spatial alignment model that will allow us to evaluate the observation likelihood \eqref{eq:likelihood-expanded}. The proposed audio-visual alignment is weakly supervised and hence it requires training data. We start by briefly describing the audio-visual training data. The training data contain pairs of audio recordings and their associated directions. Let $\widetilde{\Wmat} = \{ \Wtvect_1, \dots, \Wtvect_m, \dots \Wtvect_M\}\in\mathbb{C}^{F\times M}$ be a training dataset containing $M$ binaural vectors. Each binaural vector is extracted from its corresponding audio recording using the method described in Section~\ref{sec:single-speaker}, \ie $\Wtvect_m = (\widetilde{W}_m^1, \dots, \widetilde{W}_m^f, \dots, \widetilde{W}_m^F)$ where each entry $\widetilde{W}_m^f$ is computed with \eqref{eq:RTF-one}.

\addnote[ch:israel]{1}{Each audio sample in the training set consists of a white-noise signal that is emitted by a loudspeaker placed at different locations, e.g. Fig.~\ref{fig:dataset_setup}}. The PSD of a white-noise signal is significant at each frequency thus: $|\widetilde{W}_{m}^{f}|^2 >a>0, \forall m\in[1\dots M], \forall f\in [1 \dots F]$. A visual marker placed onto the loudspeaker allows to associate its pixel location with each sound direction, hence the $M$ source directions correspond to an equal number of pixel locations $\widetilde{\Xmat} = \{ \Xtvect_1, \dots, \Xtvect_m, \dots \Xtvect_M\}\in\mathbb{R}^{2\times M}$. To summarize, the training data consist of $M$ pairs of binaural features and associated pixel locations: $\{ \Wtvect_m, \Xtvect_m \}_{m=1}^M$.

We now consider the two sets of visual and auditory observations during the time slice $t$, namely $\Xmat_t = (\Xvect_{t,1}, \dots, \Xvect_{t,n}, \dots, \Xvect_{t,N})\in\mathbb{R}^{2\times N}$,
$\Vvect_t = (V_{t,1}, \dots V_{t,n}, \dots V_{t,N})\in \{0,1\}^{N}$, $\Ymat_t=(\Yvect_{t,1}, \dots, \Yvect_{t,k}, \dots, \Yvect_{t,K})\in\mathbb{C}^{F\times K}$ and $\Amat_t\in \{0,1\}^{F\times K}$. If person $n$, located at $\Xvect_{t,n}$, is both visible and speaks at $t$: the binaural features associated with the emitted speech signal depend on the person's location only, hence they must be similar to the binaural features of the training source emitting from the same location. This can be simply written as a nearest-neighbor search over the training-set of audio-source locations:
\begin{align}
\Xtvect_n = \argmin_{m} \| \Xvect_{t,n} - \Xtvect_m\|^2
\end{align}
and let $\Wtvect_n \in \widetilde{\Wmat}$ be the binaural feature vector associated with this location. Hence, the training pair $\{\Xtvect_n, \Wtvect_n\}\in\widetilde{\Xmat} \times \widetilde{\Wmat}$ can be associated with person $n$.
%The task of audio-visual fusion is to associate audio observations with visible persons and to estimate the speech-activity likelihood of each person. 

We choose to model that at any frequency $f\in [1\dots F]$, the likelihood of and observed binaural feature $Y_{t,k}^{f}$ follows the following complex-Gaussian mixture model (for convenience, we omit the the time index $t$)
\begin{align} \label{eq:audiovisualGMM}
P(Y_{k}^{f}|&\Thetavect^{f}) = \\ \nonumber
&\sum_{n=1}^{N} \pi_{n}^{f} \mathcal{N}_c (Y_{k}^{f} | \widetilde{W}_{n}^{f}, \sigma_{n}^{f}) + \pi_{N+1}^{f} \mathcal{N}_c (Y_{k}^{f} | 0, \sigma_{ N+1}^{f}),
\end{align}
$\text{where } \mathcal{N}_c (x|\mu,\sigma) = (\pi \sigma)\inverse \exp (-|x-\mu|^2/\sigma), \quad x\in\mathbb{C} $
is the complex-normal distribution and $\Thetavect^f$ is the set of \textit{real-valued} model parameters, namely the priors $\{\pi_{n}^{f}\}_{n=1}^{N+1}$ with $\sum_{n=1}^{N+1} \pi_{n}^{f}=1$, and the variances $\{\sigma_{n}^{f}\}_{n=1}^{N+1}$. This model states that the binaural feature $Y_{k}^{f} $ is either generated by one of the $N$ persons, located at $\Xtvect_n, 1\leq n \leq N$, hence it is an inlier generated by a complex-normal mixture model with means $ \widetilde{W}_{n}^{f}, 1\leq n \leq N$, or is emitted by an unknown sound source, hence it is 
an outlier generated by a zero-centered complex-normal distribution with a very large variance $\sigma_{ N+1}^{f} \gg \sigma_n$. 

%\addnote[ch:israel]{1}{It is important to emphasize that the $N+1$ mixture component models binaural features from all kind of unknown sound sources. Here, unknown in a sense that the sources are not currently visible or being visually tracked and hence audio-visual alignment as described above is not possible. There are three possible types of unknown sound sources. The first ones are non-person sources either in or outside the camera field of view (FoV), \eg a hi-fi equipment playing music or radio, which in general are not visully tracked. The second unknown sources are persons that are outside the FoV of the camera while speaking and hence, it is impossible to perform visual tracking. The third type of unknown sources are persons inside the FoV of the camera while speaking but the visual tracker failed to track and thus their visibility variable is set to 0. While the first two kinds are not in the interest of audio-visual speaker diarization since the task is to perform diarization for speakers only in camera FoV, the last one contributes to diarization miss errors. The diarization model proposed in this work emphasize the importance of a robust multi-target visual tracker. The discussion about robust visual tracking is beyond the scope of this work contributions.} 
%  However, it is not robust to long-term tracking failures. 

The parameter set $\Thetavect^f$ of \eqref{eq:audiovisualGMM} can be easily estimated via a simplified variant of the EM algorithm for Gaussian mixtures: the algorithm alternates between E-step that evaluates the posterior probabilities $r_{kn}^f = P(z_{k}^f = n | Y_{k}^{f})$, $z_k^f$ is assignment varaible, $z_k^f=n$ means $Y_{k}^{f}$ is generated by component $n$:
\begin{align} \label{eq:audiovisual-posteriors}
&r_{kn}^f = \begin{cases}
\frac{1}{C} \pi_{n}^{f} \mathcal{N}_c (Y_{k}^{f} | \widetilde{W}_{n}^{f}, \sigma_{n}^{f})& \text{if $1\leq n \leq N$} \\
\frac{1}{C} \pi_{N+1}^{f}\mathcal{N}_c (Y_{k}^{f} | 0, \sigma_{ N+1}^{f}) & \text{if $n = N+1$},
\end{cases} \\
&\text{where } C = \sum_{i=1}^{N} \pi_{i}^{f} \mathcal{N}_c (Y_{k}^{f} | \widetilde{W}_{i}^{f}, \sigma_{i}^{f}) +  \pi_{N+1}^{f} \mathcal{N}_c (Y_{k}^{f} | 0, \sigma_{ N+1}^{f}), \nonumber
\end{align}
% \begin{align}
% \label{eq:audiovisual-posteriors}
% &\quad \quad r_{kn}^f = \nonumber \\
% & 
% %\begin{cases}
% \begin{array}{ll}
% \frac{\pi_{n}^{f} \mathcal{N}_c (Y_{k}^{f} | \widetilde{W}_{n}^{f}, \sigma_{n}^{f}) }{\sum_{i=1}^{N} \pi_{i}^{f} \mathcal{N}_c (Y_{k}^{f} | \widetilde{W}_{i}^{f}, \sigma_{i}^{f}) +  \pi_{N+1}^{f}
% \mathcal{N}_c (Y_{k}^{f} | 0, \sigma_{ N+1}^{f})} & \text{if $1\leq n \leq N$} \\
% \frac{ \pi_{N+1}^{f}
% \mathcal{N}_c (Y_{k}^{f} | 0, \sigma_{ N+1}^{f})}{\sum_{i=1}^{N} \pi_{i}^{f} \mathcal{N}_c (Y_{k}^{f} | \widetilde{W}_{i}^{f}, \sigma_{i}^{f}) +  \pi_{N+1}^{f}
% \mathcal{N}_c (Y_{k}^{f} | 0, \sigma_{ N+1}^{f})} & \text{if $n = N+1$},
% \end{array}
% %\end{cases}
% \end{align}
and M-step that estimates the variances and the priors:
\begin{align}
\label{eq:audiovisual-variances}
\sigma_{n}^{f} &= \frac{\sum_{k=1}^K A_{k}^{f} r_{kn}^f |Y_{k}^{f} - \widetilde{W}_{n}^{f} |^2 }{\sum_{k=1}^K A_{k}^{f} r_{kn}^f } \quad \forall n, 1\leq n \leq N \\
\pi_{n}^{f} &=\frac{\sum_{k=1}^K A_{k}^{f} r_{kn}^f }{\sum_{k=1}^K A_{k}^{f}} \quad \forall n, 1\leq n \leq N+1 .
\label{eq:audiovisual-priors}
\end{align}
The algorithm can be easily initialized by setting all the priors equal to $\tfrac{1}{N+1}$ and by setting all the variances equal to a positive scalar $\sigma$. Because the component means are fixed, the algorithm converges in only a few iterations. 

Based on these results one can evaluate \eqref{eq:likelihood-expanded}, namely the speaking probability of person $n$ located at $\Xvect_n$: the probability that a visible person either speaks:
\begin{equation}
\label{eq:audiovisual-activespeaker}
P(\Umat_t | S_{t,n}=1, V_{t,n}=1) = \frac{\sum_{f=1}^F\sum_{k=1}^K A_{t,k}^{f} r_{t,kn}^f }{\sum_{f=1}^F\sum_{k=1}^K A_{t,k}^{f}},
\end{equation}
or is silent:
\begin{equation}
\label{eq:audiovisual-silentspeaker}
P(\Umat_t | S_{t,n}=0, V_{t,n}=1) = 1 - P(\Umat_t | S_{t,n}=1, V_{t,n}).
\end{equation}

%% file: audiovisualdata_v1.tex
% audio-visual datasets
\begin{figure*}[htb!]
\centering
\begin{minipage}[b]{1.99\columnwidth}
\centering
\subfloat[]{
  \begin{tikzpicture}
    \node[anchor=south west,inner sep=0] at (0,0) {\includegraphics[trim=0mm 0mm 0mm 0mm,clip,width=0.30\columnwidth,scale=0.2,keepaspectratio]{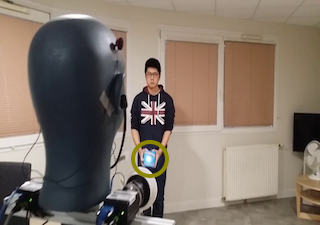}};
    \draw[green, ultra thick] (2.5,1.12) circle (0.3cm);
  \end{tikzpicture}} 
\vspace{0.25em}
%\subfloat[]{\includegraphics[trim=0mm 0mm 0mm 0mm,clip,width=0.24\columnwidth,scale=0.2,keepaspectratio]{calib_grid.png}} \vspace{0.25em}
\subfloat[\label{fig:dataset_setup_b}]{\includegraphics[trim=0mm 0mm 0mm 0mm,clip,width=0.30\columnwidth,scale=0.2,keepaspectratio]{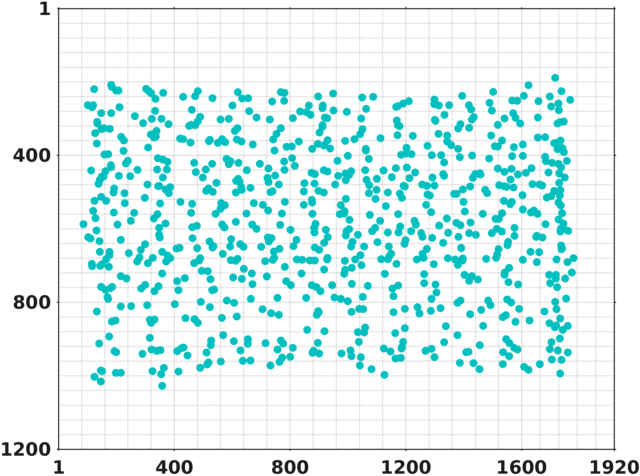}} 
\vspace{0.25em}
\subfloat[]{
  \begin{tikzpicture}
    \node[anchor=south west,inner sep=0] at (0,0) {\includegraphics[trim=0mm 0mm 0mm 0mm,clip,width=0.30\columnwidth,scale=0.2,keepaspectratio]{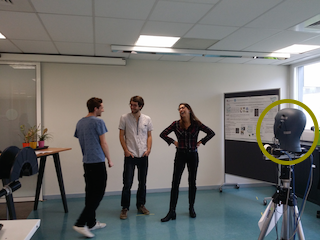}};
    \draw[green, ultra thick] (4.8,2.0) circle (0.6cm and 0.6cm);
  \end{tikzpicture}} 
\vspace{0.25em}
\end{minipage}
\caption{\label{fig:dataset_setup} The \textbf{AVDIAR} dataset is recorded with a camera-microphone setup. (a) To record the training data, a loud-speaker that emits white noise was used. A visual marker onto the loud-speaker (circled in green) allows to annotate the training data with image locations, each image location corresponds to a loud-speaker direction. (b) The image grid of loud-speaker locations used for the training data. (c) A typical \textbf{AVDIAR} scenario (the camera-microphone setup is circled in green).}
\end{figure*}

In this section we describe the audio-visual datasets that are used to test the proposed method and to compare it with several state-of-the-art methods. We start by describing a novel dataset that was purposively gathered and recorded to encompass a wide number of multiple-speaker scenarios, \eg speakers facing the camera, moving speakers, speakers looking at each other, etc. This novel dataset is referred to as \textbf{AVDIAR}.\footnote{\url{https://team.inria.fr/perception/avdiar/}}

In order to record both training and test data we used the following camera-microphone setup. A color camera is rigidly attached to an acoustic dummy head. 
The camera is a PointGrey Grasshopper3 unit equipped with a Sony Pregius IMX174 CMOS sensor of size $1.2^{\prime\prime}\times 1^{\prime\prime}$. The camera is equipped with a Kowa 6~mm wide-angle lens and it delivers 1920$\times$1200 color pixels at 25~FPS. This camera-lens setup has a horizontal~$\times$~vertical field of view of $97^\circ \times 80^\circ$.

For the audio recordings we used a binaural Senheiser MKE 2002 dummy head with two microphones plugged into its left and right ears, respectively.  
The orginal microphone signals are captured at $44100$~Hz, we have downsampled them to $16000$~Hz. The STFT, implemented with a 32~ms Hann window and 16~ms shifts between consecutive windows, is then applied separately to the left and right microphone signals. Therefore, there are 512 samples per frame and the audio frame rate is approximatively 64~FPS. Each audio frame consists of a vector composed $F=256$ Fourier coefficients covering frequencies in the range $0\;\text{Hz} - 8\;\text{kHz}$. 

The camera and the microphones are connected to a single PC and they are finely synchronized using time stamps delivered by the computer's internal clock. This audio-visual synchronization allows us to align the visual frames with the audio frames. The time index $t$ corresponds to the visual-frame index. For each $t$ we consider a spectrogram of length $K=25$ frames, or a time slice of 0.4~s, hence there is an overlap between the spectrograms corresponding to consecutive time indexes.

\begin{table*}[t!]
  \centering
  \caption{Scenarios available with the \textbf{AVDIAR} dataset.}
  \def\arraystretch{2}
  \resizebox{1.999\columnwidth}{!}{
    \begin{tabular}{|l|l|}  \hline 
      \multicolumn{1}{|c|}{\textbf{Recordings}}           &\multicolumn{1}{c|}{ \textbf{Description}} \\ \hline
      Seq01-1P-S0M1, Seq04-1P-S0M1 , Seq22-1P-S0M1	     & \pbox{1.32\columnwidth} {A single person moving randomly and alternating between speech and  silence.}\\ \hline
      Seq37-2P-S0M0, Seq43-2P-S0M0  	     		     & \pbox{1.32\columnwidth} {Two static participants taking speech turns.} \\ \hline
      Seq38-2P-S1M0, Seq40-2P-S1M0, Seq44-2P-S2M0	     & \pbox{1.32\columnwidth} {Two static participants speaking almost simultaneously, \ie there are large speech overlaps.} \\  \hline
      Seq20-2P-S1M1, Seq21-2P-S2M1		     	     & \pbox{1.32\columnwidth} {Two participants, wandering in the room and engaged in a conversation, sometime speaking simultaneously.} \\ \hline
      Seq12-3P-S2M1, Seq27-3P-S2M1	     & \pbox{1.32\columnwidth} {Three participants engaged in an informal conversation. They are moving around and sometimes they speak simultaneously.} \\ \hline
      Seq13-4P-S1M1, Seq32-4P-S1M1	     & \pbox{1.32\columnwidth} {Three to four participants engaged in a conversation. Sometimes they speak simultaneously and there are many short speech turns.}\\ 
      \hline
    \end{tabular}}
  \label{tab:seq_description}
\end{table*}

\begin{figure*}[htb]
\centering
\begin{minipage}[b]{1.999\columnwidth}
\subfloat{\includegraphics[width=0.19\columnwidth,scale=0.1,keepaspectratio]{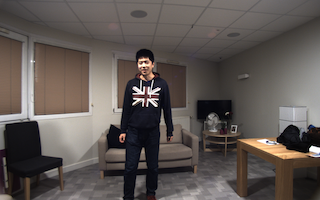}} \vspace{0.10em}
\subfloat{\includegraphics[width=0.19\columnwidth,scale=0.1,keepaspectratio]{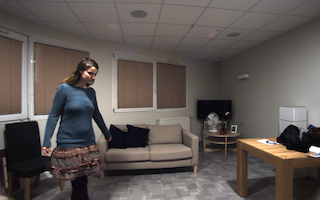}} \vspace{0.10em}
\subfloat{\includegraphics[width=0.19\columnwidth,scale=0.1,keepaspectratio]{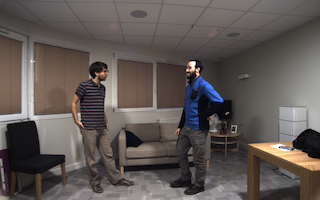}} \vspace{0.10em}
\subfloat{\includegraphics[width=0.19\columnwidth,scale=0.1,keepaspectratio]{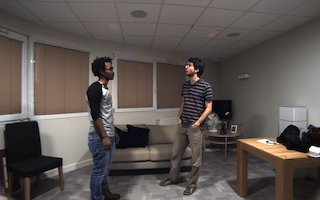}} \vspace{0.10em}
\subfloat{\includegraphics[width=0.19\columnwidth,scale=0.1,keepaspectratio]{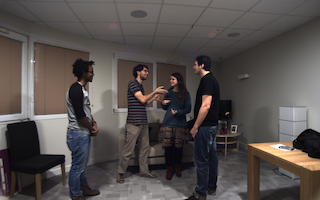}}
\end{minipage}
\par \vspace*{-1em}
\begin{minipage}[b]{1.999\columnwidth}
\subfloat{\includegraphics[width=0.19\columnwidth,scale=0.1,keepaspectratio]{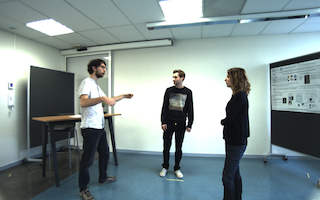}} \vspace{0.10em}
\subfloat{\includegraphics[width=0.19\columnwidth,scale=0.1,keepaspectratio]{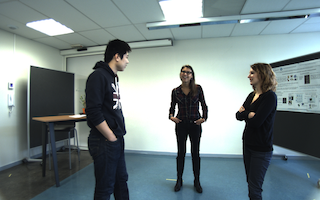}} \vspace{0.10em}
\subfloat{\includegraphics[width=0.19\columnwidth,scale=0.1,keepaspectratio]{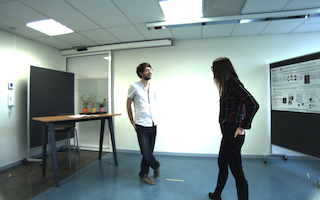}} \vspace{0.10em}
\subfloat{\includegraphics[width=0.19\columnwidth,scale=0.1,keepaspectratio]{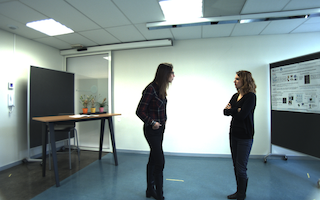}} \vspace{0.10em}
\subfloat{\includegraphics[width=0.19\columnwidth,scale=0.1,keepaspectratio]{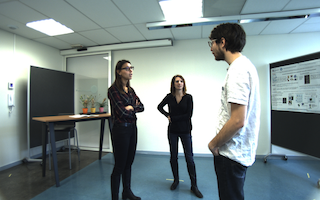}}
\end{minipage}
\par \vspace*{-1em}
\begin{minipage}[b]{1.999\columnwidth}
\subfloat{\includegraphics[width=0.19\columnwidth,scale=0.1,keepaspectratio]{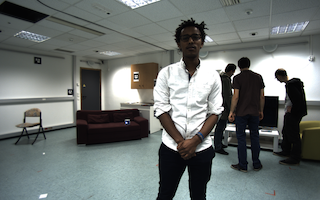}} \vspace{0.10em}
\subfloat{\includegraphics[width=0.19\columnwidth,scale=0.1,keepaspectratio]{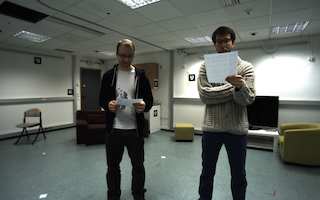}} \vspace{0.10em}
\subfloat{\includegraphics[width=0.19\columnwidth,scale=0.1,keepaspectratio]{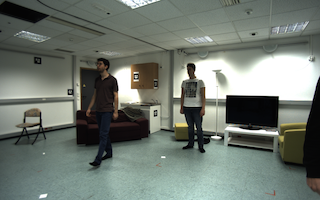}} \vspace{0.10em}
\subfloat{\includegraphics[width=0.19\columnwidth,scale=0.1,keepaspectratio]{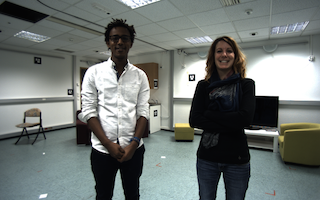}} \vspace{0.10em}
\subfloat{\includegraphics[width=0.19\columnwidth,scale=0.1,keepaspectratio]{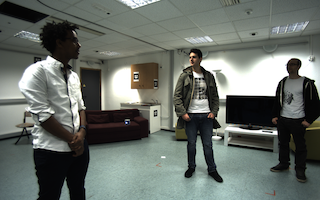}}
\end{minipage}
\caption{ \label{fig:examples_from_avdiar} Examples of scenarios in the \textbf{AVDIAR} dataset. For the sake of varying the acoustic conditions, we used three different rooms to record this dataset.}
\end{figure*} 

The training data were recorded by manually moving a loudspeaker in front of the camera-microphone unit \eg \fref{fig:dataset_setup}. A visual marker placed at the center of the loudspeaker enables recording of audio signals with their associated pixel positions in the image plane. The loudspeaker is roughly moved in two planes roughly parallel to the image plane, at 1.5~m and 2.5~m, respectively. For each plane we record $800$ positions lying on a uniform  20$\times 40$  grid that covers the entire field of view of the camera, hence there are $M=1600$ training samples. The training data consists of 1~s of white-noise (WN) signals. Using the STFT we therefore obtain two WN spectrograms of size 256$\times$64, corresponding to the left and right microphones, respectively. These two spectrograms are then used to compute binaural feature vectors, \ie Section~\ref{sec:single-speaker} (one feature vector for each loud-speaker position) and hence to build a training dataset of audio recordings and their associated image locations $\{ \widetilde{\Wmat},\widetilde{\Xmat}\} = \{ (\Wtvect_1,\Xtvect_1), \dots, (\Wtvect_m,\Xtvect_m), \dots (\Wtvect_M,\Xtvect_M)\}$, \ie Section~\ref{sec:audio-visual}.

Similarly we gathered a test dataset that contains several scenarios. Each scenario involves participants that are either static and speak or move and speak, in front of the camera-microphone unit at distance varying between 1.0~m and 3.5~m. In an attempt to record natural human-human interactions, participants were allowed to wonder around the scene and to interrupt each other while speaking. We recorded the following scenario categories, \eg \fref{fig:examples_from_avdiar}:
\begin{itemize}
\item \textit{Static participants facing the camera.} This scenario can be used to benchmark diarization methods requiring the detection of frontal faces and of facial and lip movements.
\item \textit{Static participants facing each other.} This scenario can be used to benchmark diarization methods that require static participants not necessarily facing the camera.
\item \textit{Moving participants.} This is a general-purpose scenario that can be used to benchmark diarization as well as audio-visual person tracking.
\end{itemize}

\begin{figure}[htb!]
\begin{minipage}[b]{0.99\columnwidth}
\centering
\subfloat{\includegraphics[trim=1mm 0mm 1mm 1mm,clip,width=0.230\columnwidth,scale=0.7,keepaspectratio]{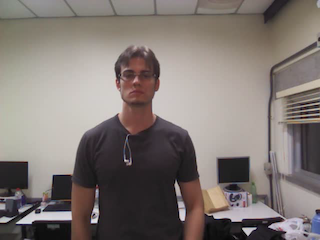}} \vspace{0.1em}
\subfloat{\includegraphics[trim=1mm 0mm 1mm 1mm,clip,width=0.230\columnwidth,scale=0.7,keepaspectratio]{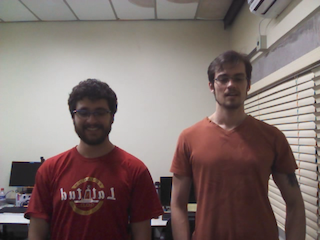}} \vspace{0.1em}
\subfloat{\includegraphics[trim=1mm 0mm 1mm 1mm,clip,width=0.230\columnwidth,scale=0.7,keepaspectratio]{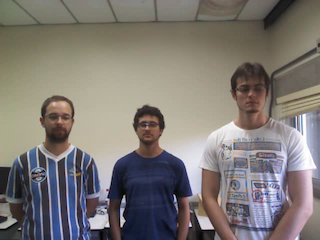}} \vspace{0.1em}
\subfloat{\includegraphics[trim=1mm 0mm 1mm 1mm,clip,width=0.230\columnwidth,scale=0.7,keepaspectratio]{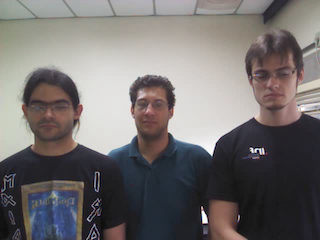}} 
\end{minipage}
\par \vspace*{-1em}
\begin{minipage}[b]{0.99\columnwidth}
\centering
\subfloat{\includegraphics[trim=1mm 1mm 1mm 1mm,clip,width=0.230\columnwidth,scale=0.6,keepaspectratio]{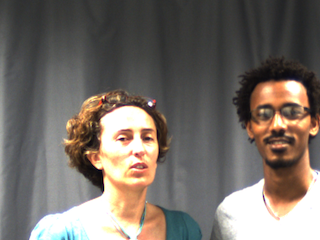}} \vspace{0.1em}
\subfloat{\includegraphics[trim=1mm 1mm 1mm 1mm,clip,width=0.230\columnwidth,scale=0.6,keepaspectratio]{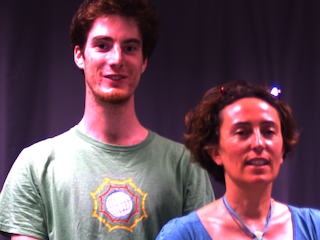}} \vspace{0.1em}
\subfloat{\includegraphics[trim=1mm 1mm 1mm 1mm,clip,width=0.230\columnwidth,scale=0.6,keepaspectratio]{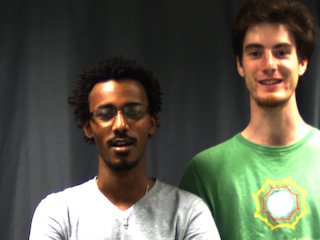}} \vspace{0.1em}
\subfloat{\includegraphics[trim=1mm 1mm 1mm 1mm,clip,width=0.230\columnwidth,scale=0.6,keepaspectratio]{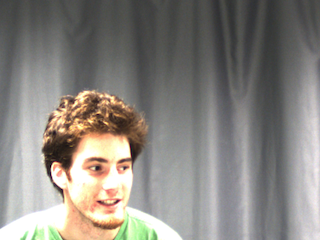}} 
\end{minipage}
\par \vspace*{-1em}
\begin{minipage}[b]{0.99\columnwidth}
\centering
\subfloat{\includegraphics[trim=1mm 1mm 1mm 1mm,clip,width=0.230\columnwidth,scale=0.6,keepaspectratio]{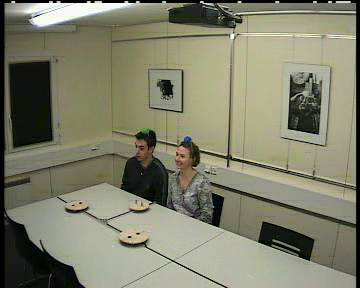}} \vspace{0.1em}
\subfloat{\includegraphics[trim=1mm 1mm 1mm 1mm,clip,width=0.230\columnwidth,scale=0.6,keepaspectratio]{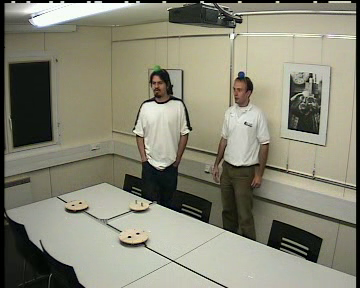}} \vspace{0.1em}
\subfloat{\includegraphics[trim=1mm 1mm 1mm 1mm,clip,width=0.230\columnwidth,scale=0.6,keepaspectratio]{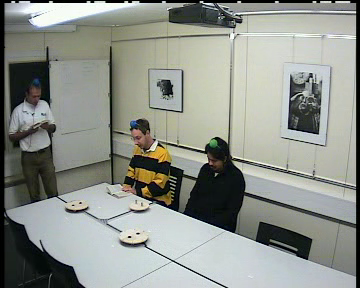}} \vspace{0.1em}
\subfloat{\includegraphics[trim=1mm 1mm 1mm 1mm,clip,width=0.230\columnwidth,scale=0.6,keepaspectratio]{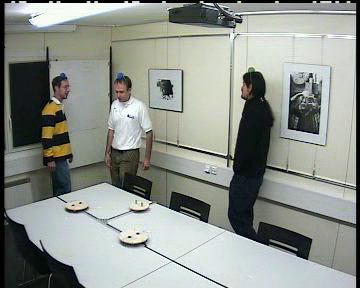}} 
\end{minipage}
\caption{\label{fig:examples_from_datasets} Example from different datasets. The \textbf{MVAD} dataset (top) contains recordings of one to three persons that always face the camera. The \textbf{AVASM} (middle) was design to benchmark audio-visual sound-source localization with two simultaneously speaking persons or with a moving speaker. The \textbf{AV16P3} dataset (bottom) contains recordings of simultaneously moving and speaking persons.}
\end{figure}

In addition to the \textbf{AVDIAR} dataset, we used three other datasets, \eg \fref{fig:examples_from_datasets}. They are briefly described as follows:
\begin{itemize}
\item The \textbf{MVAD} dataset described in \cite{minotto2015multimodal}. The visual data were recorded with a Microsoft Kinect sensor at 20~FPS,\footnote{Note that our method doesn't use the depth image available with this sensor}  and the audio signals were recorded with a linear array of omnidirectional microphones sampled at 44100 Hz. The recorded sequences are from 40~s to 60~s long  and contain one to three participants that speak in Portuguese. The speech and silence segments are 4~s to 8~s long. Since the diarization method proposed in \cite{minotto2015multimodal} requires frontal faces, the participants are facing the camera and remain static through all the recordings. 

\item The \textbf{AVASM} dataset contains both training and test recordings used to test the single and multiple speaker localization method described in \cite{deleforge2015colocalization}. The recording setup is similar to the one described above, namely a  binaural acoustic dummy head with two microphones plugged into its ears and a camera placed underneath the head. The images and the audio signals were captured at 25~FPS and 44100~Hz, respectively. The recorded sequences contain up to two participants that face the camera and speak simultaneously. In addition, the dataset has
audio-visual alignment data collected in a similar fashion as the \textbf{AVDIAR} dataset. 
 
\item The \textbf{AV16P3} dataset is designed to benchmark audio-visual tracking of several moving speakers  without taking diarization into account \cite{lathoud2004av16}. The sensor setup used for these recordings is composed of three cameras attached to the room ceiling, and two circular eight-microphone arrays. The recordings include mainly dynamic scenarios, comprising a single, as well as multiple moving speakers. In all the recordings there is a large overlap between the speaker-turns. 
\end{itemize}

These datasets contain a large variety of recorded scenarios, aimed at a wide range of application. \eg formal and informal interaction in meetings and gatherings, human-computer interaction, etc. Some of the datasets were not purposively recorded to benchmark diarization. Nevertheless they are challenging because they contain a large amount of overlap between speakers, hence they are well suited to test the limits and failures of diarization methods. Unlike recordings of formal meetings, which are composed on long single-speech segments with almost no overlap between the participants, the above datasets contain the following challenging situations \eg \tref{tab:seq_description}:
\begin{itemize}
  \item The participants do not always face the cameras, moreover, they turn their heads while they speak or listen;
  \item The participants, rather then being static, move around and hence the tasks of tracking and diarization must be finely intertwined;
  \item In informal meetings participants interrupt each other and hence not only that there is no silence between speech segments, but the speech segments overlap each other, and
  \item Participants take speech turns quite rapidly which results in short-length speech segments, which makes audio-visual temporal alignment quite challenging.
\end{itemize}

%% file: experiments_v1.tex
\subsection{Diarization Performance Measure}
To effectively benchmark our model with state-of-the art methods, we use the diarization error rate (DER) to quantitatively measure the performance:  \textit{smaller the \textbf{DER} value, better the performance}. DER is defined by the NIST-RT evaluation testbed,\footnote{\url{http://www.nist.gov/speech/tests/rt/2006-spring/}} and corresponds to the percentage of audio frames that are not correctly assigned to one or more speakers, or to none of them in case of a \textit{silent} frame. DER consists of the composition of the following measurements: 
\begin{itemize}
 \item False-alarm error, when speech has been incorrectly detected;
 \item Miss error, when a person is speaking but the method fails to detect the speech activity, and
 \item Speaker-labeling error, when a person-to-speech association does not correspond to the ground truth.
\end{itemize}
To compute DER, the \textsc{md-eval} software package of NIST-RT is used, setting the forgiveness collar to a video frame of \eg 40~ms for 25~FPS videos. 

\subsection{Diarization Algorithms and Setup}
\addnote[comparisons-1]{1}{
We compared our method with four methods: \cite{vijayasenan2012diartk}, \cite{minotto2015multimodal}, \cite{barzelay2010onsets}, and \cite{gebru:hal-01220956}. These methods are briefly explained below:
\begin{itemize}[topsep=0pt,itemsep=0.2em,partopsep=1em,parsep=0em]
\item Vijayasenan et al. \cite{vijayasenan2012diartk} (\textit{DiarTK}) use audio information only. \textit{DiarTK} allows the user to incorporate a large number of audio features. In our experiments and comparisons we used the following features: mel-frequency cepstral coefficients (MFCC), frequency-domain linear prediction (FDLP), time difference of arrival (TDOA), and modulation spectrum (MS). Notice that TDOA features can only be used with static sound-sources, hence we did not use TDOA in the case of moving speakers.
\item Minotto et al. \cite{minotto2015multimodal} learn an SVM classifier based on based on labeled audio-visual features. Sound-source localization provides horizontal sound directions which are combined with the output of a mouth tracker.
\item Barzelay et al. \cite{barzelay2010onsets} calculate audio-visual correlations based on extracting \textit{onsets} from both modalities and on aligning these onsets. 
%HERE
The method consists of detecting faces and on tracking face landmarks, such that each landmark yields a trajectory. Onset signals are then extracting from each one of these trajectory as well as from the microphone signal. These onsets are used to compare each visual trajectory with the microphone signal, and the trajectories that best match the microphone signal correspond to the active speaker. We implemented this method based on \cite{barzelay2010onsets} since there is no publicly available code. 
%The implementation of the KLT algorithm is obtained from \cite{itseez2015opencv}, while the audio onsets are detected following the detailed description of the work in \cite{bello2005tutorial}. The parameter for onsets matching were set using cross validation. 
Extensive experiments with this method revealed that frontal views of speakers are needed. Therefore, we tested this methods with all the sequences from the \textbf{MVAD} and \textbf{AVASM} datasets and on the sequences from the \textbf{AVDIAR} dataset featuring frontal images of faces.
\item Gebru et al. \cite{gebru:hal-01220956} track the active speaker, provided that participants take speech turns with no signal overlap. Therefore, whenever two persons speak simultaneously, this method extracts the \textit{dominant} speaker.
\end{itemize} 
}
\addnote[comparisons-2]{1}{
Additionally, we used the following multiple sound-source localization methods:
\begin{itemize}
\item \textit{GCC-PHAT} which detects the local maxima of the generalized cross-correlation method: we used the implementation from the BSS Locate Toolbox \cite{blandin2012multi}.
\item \textit{TREM} which considers a regular grid of source locations and selects the most probable locations based on maximum likelihood: we used the Matlab code provided by the authors, \cite{dorfan2015tree}.  
\end{itemize}
GCC-PHAT and TREM were used in conjunction with the proposed diarization method using the \textbf{AVDIAR} dataset as well as the \textbf{MVAD} and \textbf{AV3P16} datasets.
}

\subsection{Results and Discussion}
%We now present and discuss the results obtained on each dataset, and conclude with an overall discussion at the end of this section. 

The results obtained with the \textbf{MVAD}, \textrm{AVASM}, \textbf{AV16P3} and \textbf{AVDIAR}  datasets are summarized in \tref{tab:der_mvad}, \tref{tab:der_avasm}, \tref{tab:der_av16p3} and \tref{tab:der_avdiar}, respectively. 

\addnote[discussion-1]{1}{
Overall, it can be noticed that the method of \cite{barzelay2010onsets} is the least performing method. As explained above this method is based on detecting signal onsets in the two modalities and on finding cross-modal correlations based on onset coincidence. Unfortunately, the visual onsets are unable to properly capture complex speech dynamics. The \textit{DiarTK} method of \cite{vijayasenan2012diartk} is the second least performing method. This is mainly due to the fact that this method is designed to rely on long speech segments with almost no overlap between consecutive segments. Whenever several speech signals overlap, it is very difficult to extract reliable information with MFCC features, since the latter are designed to characterize clean speech. \textit{DiarTK} is based on clustering MFCC features using a Gaussian mixture model. Consider, for example, MFCC feature vectors of dimension 19, extracted from 20~ms-long audio frames, and a GMM with diagonal covariance matrices. If it is assumed that a minimum of 50 samples are needed to properly estimate the GMM parameters, speech segments of at least 50$\times$19$\times$20~ms, or 19~s, are needed. Therefore it is not surprising that \textit{DiarTK} performs poorly on all these datasets. %As already noted above, audio-visual diarization methods relying on temporal cross-modal correlation, \eg \cite{barzelay2010onsets,garau2010audio,noulas2012multimodal}, suffer from the same limitation.
}

\addnote[discussion-2]{1}{
Table~\ref{tab:der_mvad} shows that the method of \cite{minotto2015multimodal} performs much better than \textit{DiarTK}. This is not surprising, since the speech turns taken by the participants in the \textbf{MVAD} dataset are very brief. Minotto et al. \cite{minotto2015multimodal} use a combination of visual features extracted form frontal views of faces (lip movements) and audio features (speech-source directions) to train an SVM classifier. The method fails whenever the participants do not face the camera, \eg sequences \textit{Two12}, {Two13} and \textit{Two14}, where participants purposely occlude their faces several times throughout the recordings. The method proposed in this paper in combination with \textit{TREM} achieves the best results on almost all the tested scenarios. This is due to the fact that the audio-visual fusion method is capable of associating very short speech segments with one or several participants. However, the performance of our method, with either \textit{TREM} or \textit{GCC-PHAT}, drops down as the number of people increases. This is mainly due to the limited resolution of multiple sound-source localization algorithms (of the order of $10^\circ$ horizontally) and thus, it makes it difficult to disambiguate two nearby speaking/silent persons. Notice that tracking the identity of the participants is performed by visual tracking, which is a trivial task for most of these recordings, since participants are mostly static.
% This is because of the limitation of both SSL algorithms is that it is very difficult to find clear peaks in $\phi(\tau)$ function that is good enough to disambiguate nearby speaking persons. The performance of our model slightly drops when two speakers get closer to each other, \eg sequences \textit{Two13}, \textit{Three2}, and \textit{Three3}. 
}

\addnote[discussion-3]{1}{
Table~\ref{tab:der_avasm} shows the results obtained with the \textbf{AVASM} dataset. In these recordings the participants speak simultaneously, with the exception of the \textit{Moving-Speaker-01} recording. We do not report results obtained with \textbf{DiarTK} since this method yields non-meaningful performance with this dataset. The proposed method performs reasonable well in the presence of simultaneously speaking persons. 
}

\addnote[discussion-4]{1}{
Table \ref{tab:der_av16p3} shows results obtained with the \textbf{AV16P3} dataset. As with the \textbf{AVASM} dataset we were unable to obtain meaningful results with the \textbf{DiarTK} method. As expected the proposed method has the same performance as \cite{gebru:hal-01220956} in the presence of a single active speaker, \eg \textit{seq11-1p-0100} and \textit{seq15-1p-0111}. Nevertheless, the performance of \cite{gebru:hal-01220956} rapidly degrades in the presence of two and three persons speaking almost simultaneously. Notice that this dataset was recorded to benchmark audio-visual tracking, not diarization.
}
\begin{table}[htb!]
\caption{DER scores obtained with MVAD dataset (\%). }
\centering
\def\arraystretch{1.5}
\resizebox{0.9\columnwidth}{!}{
\begin{tabular}{l*{6}{c}}
\toprule
Sequence     &  \textbf{DiarTK} \cite{vijayasenan2012diartk} & \cite{minotto2015multimodal}  &  \cite{barzelay2010onsets}  & \cite{gebru:hal-01220956} & \specialcell{Proposed with\\ \textit{TREM} \cite{dorfan2015tree}} & \specialcell{Proposed with\\ \textit{GCC-PHAT} \cite{blandin2012multi}}
\\
\midrule
One7		     & 21.16 & 8.66     & 89.90 & 5.82  & 0.91   & 1.06 \\
One8		     & 20.07 & 7.11     & 98.10 & 4.92  & 1.02   & 1.81 \\
One9		     & 22.79 & 9.02     & 94.60 & 13.66 & 0.98   & 1.58 \\
Two1		     & 23.50 & 6.81     & 94.90 & 16.79 & 2.87   & 26.00 \\
Two2		     & 30.22 & 7.32     & 90.60 & 23.49 & 3.13   & 13.70 \\
Two3		     & 25.95 & 7.92     & 94.50 & 25.75 & 8.30   & 20.88 \\
Two4		     & 25.24 & 6.91     & 84.10 & 20.23 & 0.16   & 11.20 \\
Two5		     & 25.96 & 8.30     & 90.80 & 25.02 & 4.50   & 29.67 \\
Two6		     & 29.13 & 6.89     & 96.70 & 16.89 & 6.11   & 23.57 \\
Two9		     & 30.71 & 11.95    & 96.90 & 15.59 & 2.42   & 34.28 \\
Two10		     & 25.32 & 8.30     & 95.50 & 21.04 & 3.27   & 15.15 \\
Two11		     & 27.75 & 6.12     & 84.60 & 21.22 & 6.89   & 18.05 \\
Two12		     & 45.06 & 24.60    & 80.40 & 39.79 & 12.00  & 34.60 \\
Two13		     & 49.23 & 27.38    & 64.10 & 25.11 & 14.49  & 48.70 \\
Two14		     & 27.16 & 28.81    & 81.10 & 25.75 & 6.43   & 59.10 \\
Three1		   & 27.71 & 9.10     & 95.80 & 47.56 & 6.17   & 52.63 \\
Three2		   & 27.71 & 9.10     & 89.20 & 49.15 & 13.46  & 49.66 \\
Three3		   & 29.41 & 5.93     & 91.50 & 47.78 & 13.57  & 49.09 \\
Three6		   & 36.36 & 8.92     & 79.70 & 40.92 & 12.89  & 37.78 \\
Three7		   & 36.24 & 14.51    & 86.20 & 47.35 & 11.74  & 40.40 \\
\bottomrule
\textbf{Average}          & 29.33 & 11.18    & 89.96 & 26.69 & 6.57   & 28.45\\
\bottomrule
\end{tabular} }
\label{tab:der_mvad}
\end{table}

\begin{table}[htb!]
  \caption{DER scores obtained with AVASM dataset (\%).}
  \centering
  \def\arraystretch{1.5}
  \resizebox{0.9\columnwidth}{!}{
    \begin{tabular}{l*{6}{c}}
      \toprule
      Sequence             &\cite{barzelay2010onsets}  & \cite{gebru:hal-01220956} & \specialcell{Proposed with\\ \textit{TREM} \cite{dorfan2015tree}} & \specialcell{Proposed with \\ \textit{GCC-PHAT} \cite{blandin2012multi}} & Proposed \\
      \midrule                                             
      Moving-Speaker-01	   	   & 95.04 & 6.26  & 21.84 & 17.24 & 6.26  \\
      Two-Speaker-01		   & 70.20 & 24.11 & 34.41 & 44.42 & 2.96  \\
      Two-Speaker-02		   & 80.30 & 26.98 & 32.52 & 47.30 & 7.33  \\
      Two-Speaker-03		   & 74.20 & 35.26 & 46.77 & 47.77 & 13.78 \\
      \bottomrule
      \textbf{Average}		   & 79.94 & 23.15 & 33.89 & 39.18 & 7.58  \\
      \bottomrule		  
    \end{tabular} }
  \label{tab:der_avasm}
\end{table}

\begin{table}[!htb]
  \caption{DER scores obtained with AV16P3 dataset (\%).}
  \centering
  \def\arraystretch{1.5}
  \resizebox{0.7\columnwidth}{!}{
    \begin{tabular}{l*{4}{c}}
      \toprule
      Sequence           & \cite{gebru:hal-01220956}  & \specialcell{Proposed with\\ \textit{TREM} \cite{dorfan2015tree}} & \specialcell{Proposed with \\ \textit{GCC-PHAT}  \cite{blandin2012multi}} \\
      \midrule
      seq11-1p-0100	& 3.50  & 3.25  & 12.18 \\
      seq15-1p-0111	& 3.29  & 3.29  & 25.28 \\
      seq18-2p-0101	& 23.54 & 7.69  & 9.13 \\
      seq24-2p-0111	& 43.21 & 17.39 & 46.50 \\
      seq40-3p-1111	& 26.98 & 8.51  & 21.03 \\
      \bottomrule
      \textbf{Average} 	& 20.04 & 8.02  & 22.82 \\
      \bottomrule
    \end{tabular} }
  \label{tab:der_av16p3}
\end{table}

\begin{table}[!htb]
\caption{DER scores obtained with AVDIAR dataset (\%). }
\centering
\def\arraystretch{1.6}
\resizebox{0.93\columnwidth}{!}{
\begin{tabular}{l*{6}{c}}
\toprule
Sequence             & \textit{DiarTK} \cite{vijayasenan2012diartk}   & \cite{barzelay2010onsets}  & \cite{gebru:hal-01220956}  & \specialcell{Proposed with\\ \textit{TREM} \cite{dorfan2015tree}} & \specialcell{Proposed with \\ \textit{GCC-PHAT} \cite{blandin2012multi}} & Proposed\\
\midrule                           
Seq01-1P-S0M1	     & 43.19  &  -     & 14.36   & 61.15   & 72.06   & 3.32   \\
Seq04-1P-S0M1	     & 32.62  &  -     & 14.21   & 71.34   & 68.84   & 9.44   \\
Seq22-1P-S0M1	     & 23.53  &  -     & 2.76    & 56.75   & 67.36   & 4.93   \\
Seq37-2P-S0M0	     & 12.95  & 34.70  & 1.67    & 41.02   & 45.90   & 2.15   \\
Seq43-2P-S0M0        & 76.10  & 79.90  & 23.25   & 46.81   & 56.90   & 6.74   \\
Seq38-2P-S1M0	     & 47.31  & 59.20  & 43.01   & 47.89   & 47.38   & 16.07  \\
Seq40-2P-S1M0        & 48.74  & 51.80  & 31.14   & 42.20   & 44.62   & 14.12  \\
Seq20-2P-S1M1	     & 43.58  &  -     & 51.78   & 58.82   & 59.38   & 35.46  \\
Seq21-2P-S2M1	     & 32.22  &  -     & 27.58   & 63.03   & 60.52   & 20.93  \\
Seq44-2P-S2M0	     & 54.47  &  -     & 44.98   & 55.69   & 51.0    & 5.46   \\
Seq12-3P-S2M1	     & 63.67  &  -     & 26.55   & 28.30   & 61.20   & 17.32  \\
Seq27-3P-S2M1	     & 46.05  &  -     & 20.84   & 47.40   & 68.79   & 18.72  \\
Seq13-4P-S1M1	     & 47.56  &  -     & 43.57   & 28.49   & 48.23   & 29.62  \\
Seq32-4P-S1M1	     & 41.51  &  -     & 43.26   & 33.36   & 71.98   & 30.20  \\
\bottomrule                                                
\textbf{Average}     & 43.82  & 56.40  & 27.78   & 48.72   & 58.87   & 15.32  \\
\bottomrule
\end{tabular} }
\label{tab:der_avdiar}
\end{table}

\input{avdiar_results}

\addnote[discussion-5]{1}{
Table~\ref{tab:der_avdiar} shows the results obtained with the \textbf{AVDIAR} dataset. The content of each scenario is briefly described in \tref{tab:seq_description}. The proposed method outperforms all other methods. It is also interesting to notice that our full method performs better than with either \textit{TREM} or \textit{GCC-PHAT}. This is due to the robust semi-supervised audio-visual association method proposed above. \fref{fig:result_seq32}, \fref{fig:result_seq12}, and \fref{fig:result_seq01} illustrate the audio-visual diarization results obtained by our method with three scenarios}.\footnote{
Videos illustrating the performance of the proposed method using these scenarios are available at \url{https://team.inria.fr/perception/avdiarization/}.}

% Table~\ref{tab:der_avdiar} shows the results obtained with the \textbf{AVDIAR} dataset. Unfortunately, we could not benchmark the methods described in \cite{garau2010audio,noulas2012multimodal,minotto2015multimodal} because there is no publicly available software for these methods. The content of each scenario is briefly described in \tref{tab:seq_description}. The proposed method outperforms the DiarTK audio-only baseline method as well as the speech-turn detection method of \cite{gebru:hal-01220956}. \fref{fig:result_seq32}, \fref{fig:result_seq12}, and \fref{fig:result_seq01} illustrate the audio-visual diarization results obtained with three scenarios.
% Videos illustrating the performance of the proposed method using these scenarios are available at \url{https://team.inria.fr/perception/avdiarization/}.

%% file: avdiar_results.tex
\begin{figure*}[htp]
\begin{center}
\begin{minipage}[b]{1.5\columnwidth}
\subfloat{\includegraphics[trim=0mm 0mm 0mm 0mm,clip,width=0.24\columnwidth,scale=0.1,keepaspectratio]{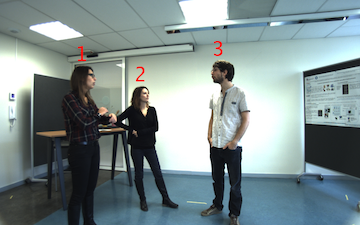}} \vspace{0.25em}
\subfloat{\includegraphics[trim=0mm 0mm 0mm 0mm,clip,width=0.24\columnwidth,scale=0.1,keepaspectratio]{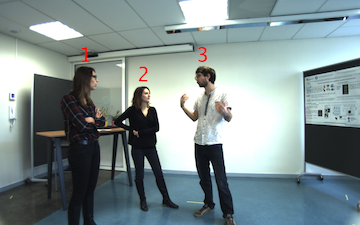}} \vspace{0.25em}
\subfloat{\includegraphics[trim=0mm 0mm 0mm 0mm,clip,width=0.24\columnwidth,scale=0.1,keepaspectratio]{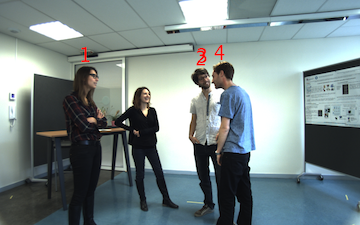}} \vspace{0.25em}
\subfloat{\includegraphics[trim=0mm 0mm 0mm 0mm,clip,width=0.24\columnwidth,scale=0.1,keepaspectratio]{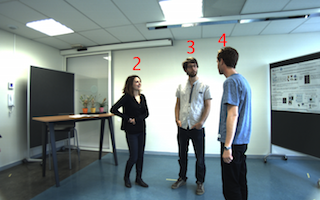}} \vspace{0.25em}
\end{minipage}
\begin{minipage}[b]{1.5\columnwidth}
 \subfloat{\includegraphics[trim=0mm 5mm 0mm 2mm,clip,width=0.99\columnwidth,scale=0.1,keepaspectratio]{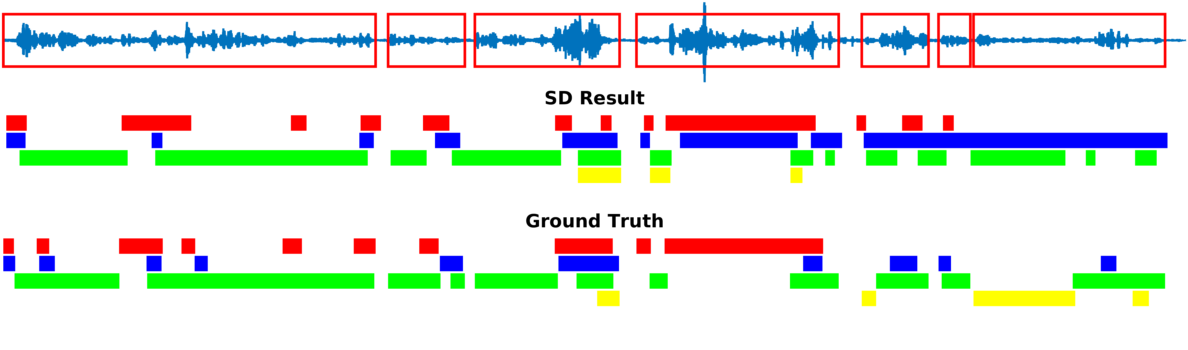}} \vspace{0.25em}
\end{minipage}
\caption{\label{fig:result_seq32} Results obtained on sequence Seq32-4P-S1M1. Visual tracking results (first row). The raw audio signal delivered by the left microphone and the speech activity region is marked with red rectangles (second row). Speaker diarization result (third row) illustrated with a color diagram: each color corresponds to the speaking activity of a different person. Annotated ground-truth diarization (fourth row).}
\end{center}
\end{figure*}

\begin{figure*}[htp]
\begin{center}
\begin{minipage}[b]{1.5\columnwidth}
\subfloat{\includegraphics[trim=0mm 0mm 0mm 0mm,clip,width=0.24\columnwidth,scale=0.1,keepaspectratio]{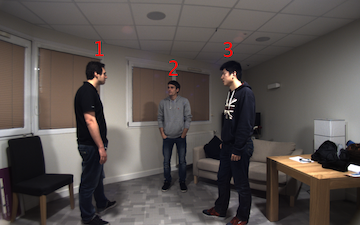}} \vspace{0.25em}
\subfloat{\includegraphics[trim=0mm 0mm 0mm 0mm,clip,width=0.24\columnwidth,scale=0.1,keepaspectratio]{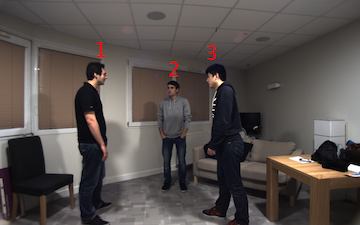}} \vspace{0.25em}
\subfloat{\includegraphics[trim=0mm 0mm 0mm 0mm,clip,width=0.24\columnwidth,scale=0.1,keepaspectratio]{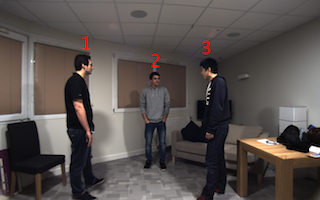}} \vspace{0.25em}
\subfloat{\includegraphics[trim=0mm 0mm 0mm 0mm,clip,width=0.24\columnwidth,scale=0.1,keepaspectratio]{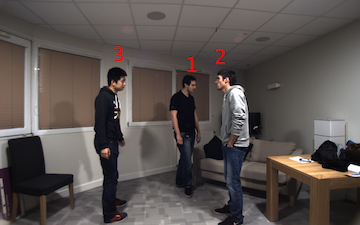}} \vspace{0.25em}
\end{minipage}
\begin{minipage}[b]{1.5\columnwidth}
 \subfloat{\includegraphics[trim=0mm 10mm 0mm 5mm,clip,width=0.99\columnwidth,scale=0.1,keepaspectratio]{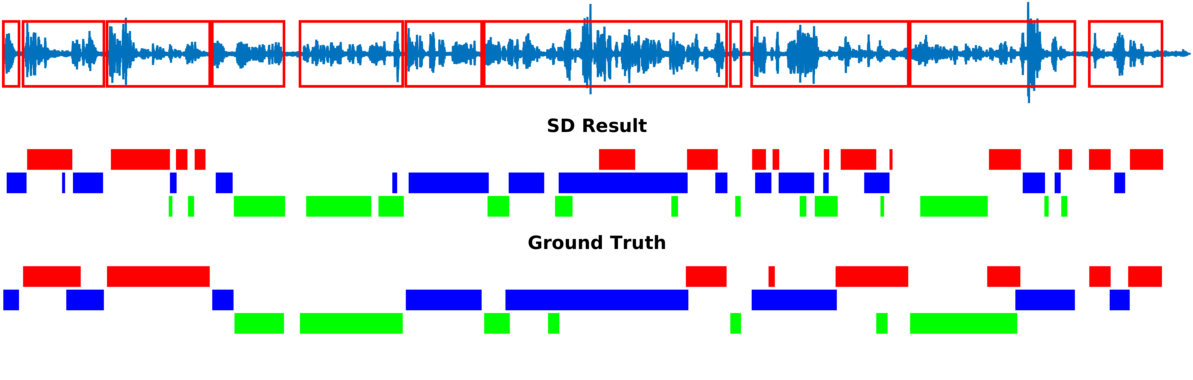}} \vspace{0.25em}
\end{minipage}
\caption{\label{fig:result_seq12} Results on sequence Seq12-3P-S2M1.}
\end{center}
\end{figure*}

\begin{figure*}[htp]
\begin{center}
\begin{minipage}[b]{1.5\columnwidth}
\subfloat{\includegraphics[trim=0mm 0mm 0mm 0mm,clip,width=0.24\columnwidth,scale=0.1,keepaspectratio]{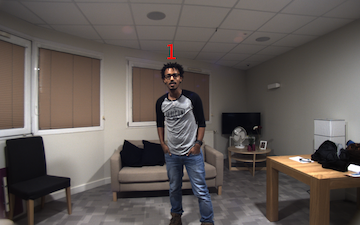}} \vspace{0.25em}
\subfloat{\includegraphics[trim=0mm 0mm 0mm 0mm,clip,width=0.24\columnwidth,scale=0.1,keepaspectratio]{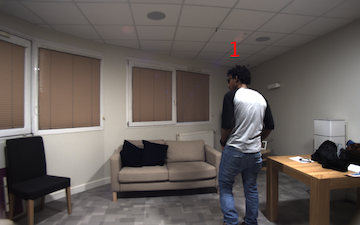}} \vspace{0.25em}
\subfloat{\includegraphics[trim=0mm 0mm 0mm 0mm,clip,width=0.24\columnwidth,scale=0.1,keepaspectratio]{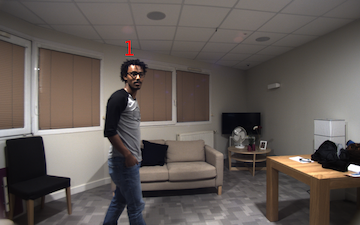}} \vspace{0.25em}
\subfloat{\includegraphics[trim=0mm 0mm 0mm 0mm,clip,width=0.24\columnwidth,scale=0.1,keepaspectratio]{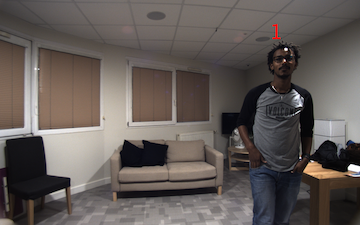}} \vspace{0.25em}
\end{minipage}
\begin{minipage}[b]{1.5\columnwidth}
 \subfloat{\includegraphics[trim=0mm 0mm 0mm 7mm,clip,width=0.99\columnwidth,scale=0.1,keepaspectratio]{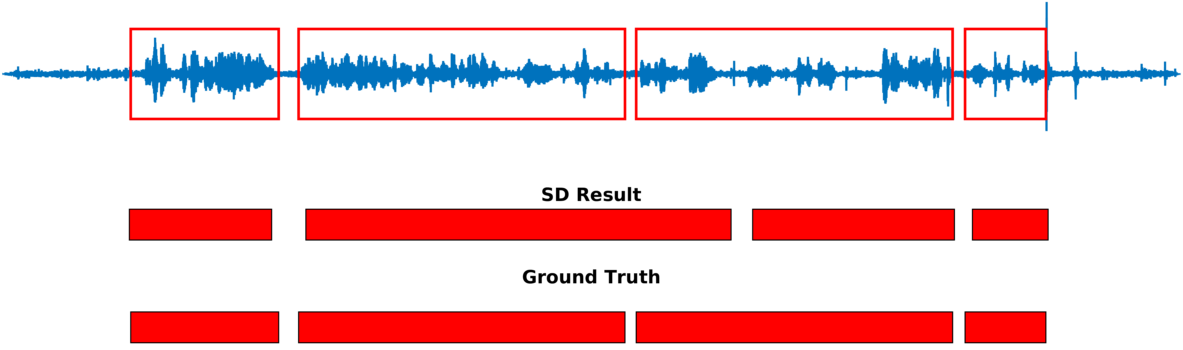}} \vspace{0.25em}
\end{minipage}
\caption{\label{fig:result_seq01} Results on sequence Seq01-1P-S0M1.}
\end{center}
\end{figure*}